\documentclass[sigconf]{acmart}

\usepackage{booktabs} 

\usepackage{amsmath}
\usepackage{epsfig}
\usepackage{booktabs} 
\usepackage{amssymb}
\usepackage{xspace}
\usepackage{algorithm}
\usepackage{algorithmic}
\usepackage{color}
\usepackage{subfigure} 
\usepackage{textcomp}
\usepackage{url}
\newcommand{\ie}{i.\nolinebreak[4]\hspace{0.01em}\nolinebreak[4]e.\@\xspace}
\newcommand{\eg}{e.\nolinebreak[4]\hspace{0.01em}\nolinebreak[4]g.\@\xspace}
\newcommand{\etal}{\emph{et al.}\xspace}
\newcommand{\Reals}{{\mathbb R}}
\newcommand{\Nats}{{\mathbb N}}
\newcommand{\tdim}{d}
\newcommand{\tsize}{n}

\newcommand{\trainset}{T}

\newcommand{\tree}{\mathcal{T}}

\newcommand{\forestNTrees}{M}
\newcommand{\forestTreeIndex}{m}
\newcommand{\leafbucketsize}{M}
\newcommand{\chunksize}{C}
\newcommand{\nbottomtrees}{n_{\text{b}}}
\newcommand{\ntoptrees}{n_{\text{t}}} 
\newcommand{\topsubsetsize}{R}
\newcommand{\topsubset}{S}
\newcommand{\numtoptreeleaves}{N}

\renewcommand{\vec}[1]{\mathbf{#1}}

\newcommand{\Yspace}{\mathcal{Y}}

\DeclareMathOperator*{\argmax}{argmax}

\newcommand{\susy}{\texttt{susy}\xspace}
\newcommand{\higgs}{\texttt{higgs}\xspace}

\newcommand{\covtype}{\texttt{covtype}\xspace}
\newcommand{\landsat}{\texttt{landsat-osm}\xspace}

\newcommand{\sklearn}{\texttt{sklearn}\xspace}
\newcommand{\subsetwood}{\texttt{subsets}\xspace}
\newcommand{\woody}{\texttt{woody}\xspace}

\newcommand{\htwo}{\texttt{h2o}\xspace}
\newcommand{\HTwoO}{$\mbox{H}_2 \mbox{O}$\xspace}

\usepackage{tikz}
\usetikzlibrary{shapes,snakes,shadows}
\newcommand*{\mycolorbox}[1]{%
\tikzstyle{mybox} = [draw=black, rectangle, inner sep=1pt,thin, fill=white]
\tikzstyle{fancytitle} = [fill=white, text=black]
\begin{tikzpicture}
\node [mybox, drop shadow={opacity=0.3,shadow xshift=.3ex, shadow yshift=-.3ex}] (box){%
     #1
};
\end{tikzpicture}%
}

\renewcommand\footnotetextcopyrightpermission[1]{}
\begin{document}
\title{Training Big Random Forests with Little Resources}

\author{Fabian Gieseke}
\affiliation{%
  \institution{Department of Computer Science\\University of Copenhagen}
  \streetaddress{Sigurdsgade 41}
  \city{Copenhagen}
  \state{Denmark}
  \postcode{2200}
}
\email{fabian.gieseke@di.ku.dk}

\author{Christian Igel}
\affiliation{%
  \institution{Department of Computer Science\\University of Copenhagen}
  \streetaddress{Sigurdsgade 41}
  \city{Copenhagen}
  \state{Denmark}
  \postcode{2200}
}
\email{igel@di.ku.dk}
%
%
%
%
%

\renewcommand{\shortauthors}{Gieseke and Igel}

\begin{abstract}
Without access to large compute clusters, building random forests on large datasets is still a challenging problem. This is, in particular, the case if fully-grown trees are desired. We propose a simple yet effective framework that allows to efficiently construct ensembles of huge trees for hundreds of millions or even billions of training instances using a cheap desktop computer with commodity hardware. The basic idea is to consider a multi-level construction scheme, which builds top trees for small random subsets of the available data and which subsequently distributes all training instances to the top trees' leaves for further processing. While being conceptually simple, the overall efficiency crucially depends on the particular implementation of the different phases. The practical merits of our approach are demonstrated using dense datasets with hundreds of millions of training instances.
\end{abstract}

%
%



\maketitle

\section{Introduction}
While large amounts of training data offer the opportunity to improve the quality of data mining models, they can also render both the generation and the
application of such models very challenging. 
Ideally, one would like to take all available data into account
during the training phase: Using more (i.i.d.) data can---in expectation---not
decrease the generalization performance and improves theoretical
generalization guarantees. 
Further, when searching for very rare patterns, 
ignoring parts of the training data or using subsampling strategies can lead to
suboptimal models (simply randomly discarding training instances from the
``negative'' class can make it difficult to define the
decision boundary around the rare ``positive'' instances).
Such learning scenarios often occur in practice, for example in astronomy or remote sensing. 

Ensemble methods are among the most successful models in data mining
and machine learning~\cite{HastieTF2009,Murphy2012}. This is
especially the case for random forests~\cite{Breiman2001}, which often
yield very competitive accuracies while being, at the same time,
conceptually simple and resilient against small changes of their hyperparameters~\cite{delgado:14}. Random forests have been
extended and modified in various  ways, \eg, to fit the requirements of special application domains or to build the involved trees in a parallel or distributed fashion in the context of large-scale learning scenarios. 
Ideally, one would like to build forests consisting of hundreds or even thousands of trees. However, depending on the data at hand, the construction of such forests can become extremely time- and memory-intensive.

For this reason, there has been a growing interest in developing frameworks and techniques that reduce the practical runtime for both the construction and the application of random forests. A popular line of research focuses on the construction of such tree ensembles in a parallel or distributed way making use of many individual compute nodes (\eg, by constructing one tree per compute node). While this can significantly reduce the practical runtime, such frameworks naturally require expensive distributed computing environments. Further, the efficient construction of a single tree might cause problems in case the dataset or the tree becomes too large to fit into the main memory of a single system.

In this work, we propose a simple yet effective construction scheme for building random forests with fully-grown trees at large scale. The main idea is to build each of the involved trees in three phases: Starting with a top tree built from a small random subset of the data, one subsequently distributes all training instances to the leaves of that tree. For each leaf subset, one builds one or more associated bottom tree(s). The intermediate leaf subsets can be stored on hard disk and, subsequently, handled individually in parallel. Hence, by using such top trees, one essentially obtains a partition of the data into much smaller and, hence, manageable subsets. Our experimental evaluation shows that our implementation can efficiently handle learning scenarios with hundreds of millions of training instances using systems with both limited computational and memory resources. To the best of the authors' knowledge, no other publicly available implementation exists that can handle datasets of this size without resorting to compute clusters.


\section{Background}
\label{section:background}
We start by providing the background related to the construction of large-scale random forests.
Let $\trainset=\{(\vec{x}_1,y_1), \ldots, (\vec{x}_\tsize, y_\tsize)\} \subset \Reals^\tdim \times \Yspace$ be a set of training patterns with $\Yspace=\Reals$ for regression and $\Yspace=\{c_1, \ldots, c_k\}$ for classification. A random forest is an ensemble of $\forestNTrees$ trees whose prediction $f(\vec{x})$ for a new pattern $\vec{x} \in \Reals^\tdim$ is based on a combination of the predictions $f_m(\vec{x})$ made by the individual trees $f_1, \ldots, f_\forestNTrees$, \ie, \begin{equation}
 f(\vec{x}) = \mathcal{C} \left( f_1(\vec{x}), \ldots, f_\forestNTrees(\vec{x}) \right),
\end{equation}
where $\mathcal{C}:\Reals^\tdim \rightarrow \Reals$ depends on the
learning scenario. For regression tasks, a common choice is $\mathcal{C} \left( f_1(\vec{x}),
  \ldots, f_\forestNTrees(\vec{x}) \right) = \frac{1}{\forestNTrees} \sum_{\forestTreeIndex=1}^\forestNTrees f_\forestTreeIndex(\vec{x})$, whereas $\mathcal{C} \left( f_1(\vec{x}),
  \ldots, f_\forestNTrees(\vec{x}) \right)
=\operatorname{argmax}_{c\in\Yspace}|\{i\,|\,f_i(\vec{x})=c\}|$ is the standard choice for classification
tasks~\cite{HastieTF2009,Murphy2012,Breiman2001}. 

\begin{figure}
\centering
 \subfigure[Tree 1]{%
  \resizebox{0.28\columnwidth}{!}{\includegraphics{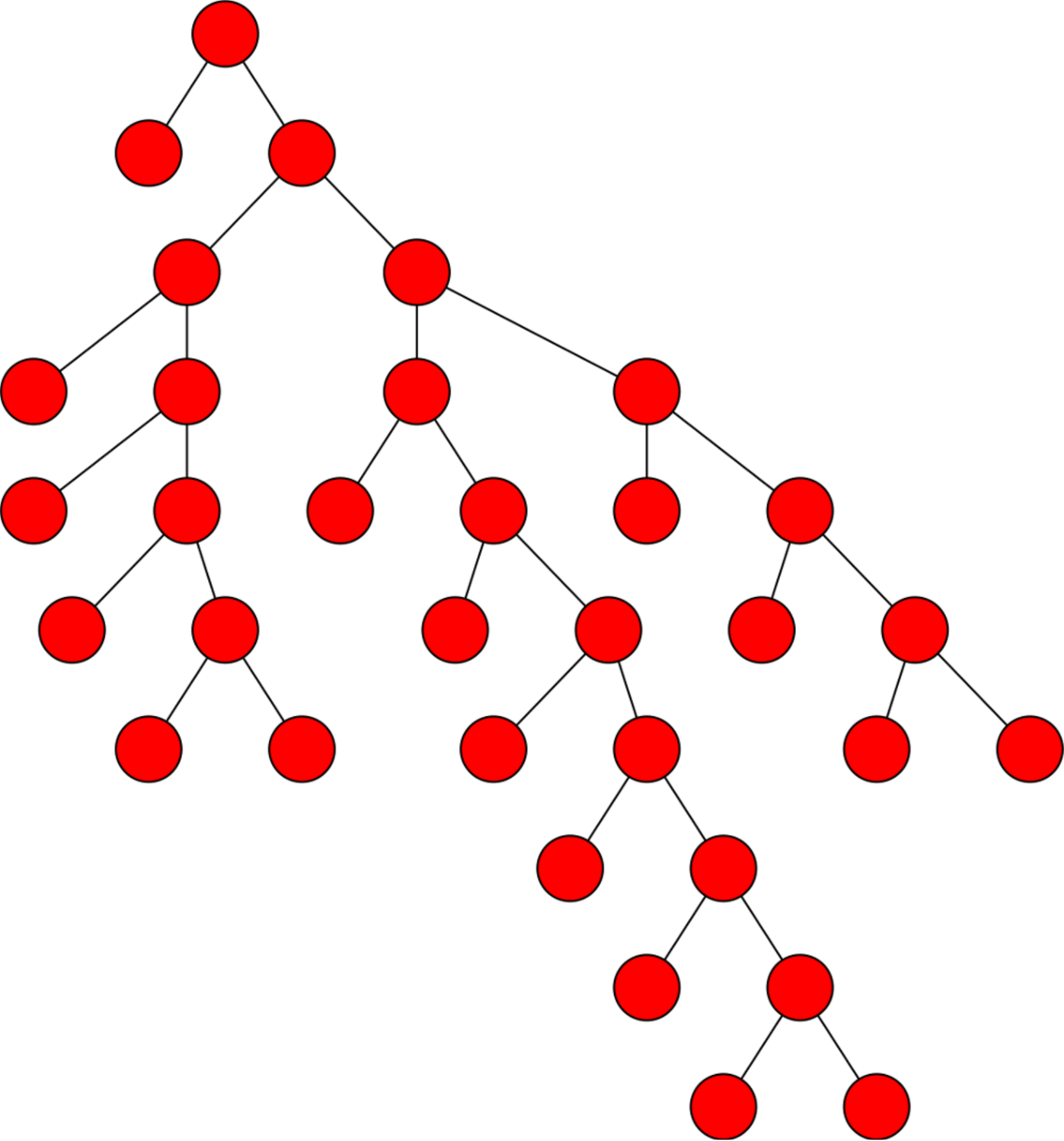}}
 }   	
 \hfill    
 \subfigure[Tree 2]{%
  \resizebox{0.28\columnwidth}{!}{\includegraphics{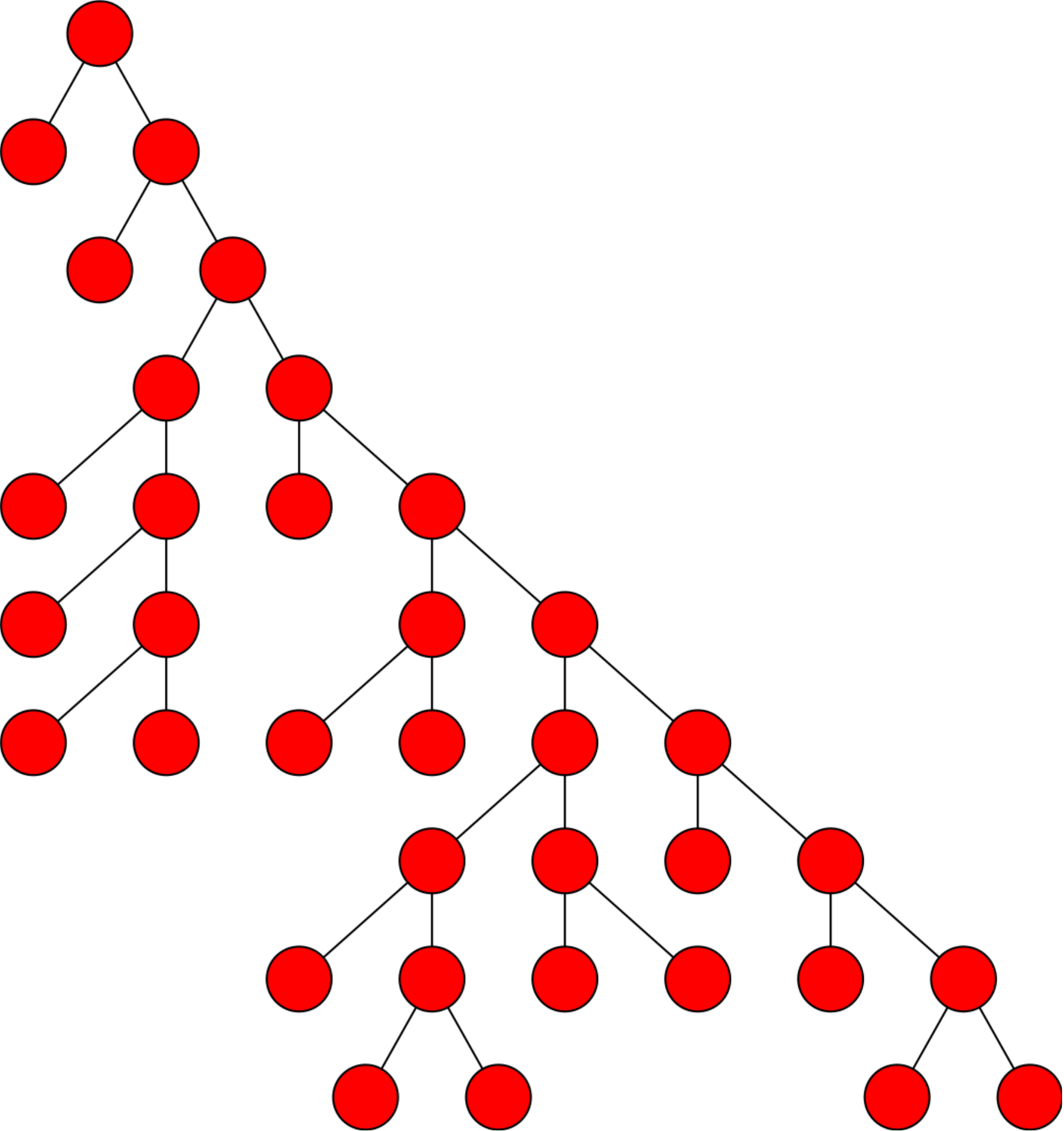}}
 }   	
 \hfill
 \subfigure[Tree 3]{%
  \resizebox{0.28\columnwidth}{!}{\includegraphics{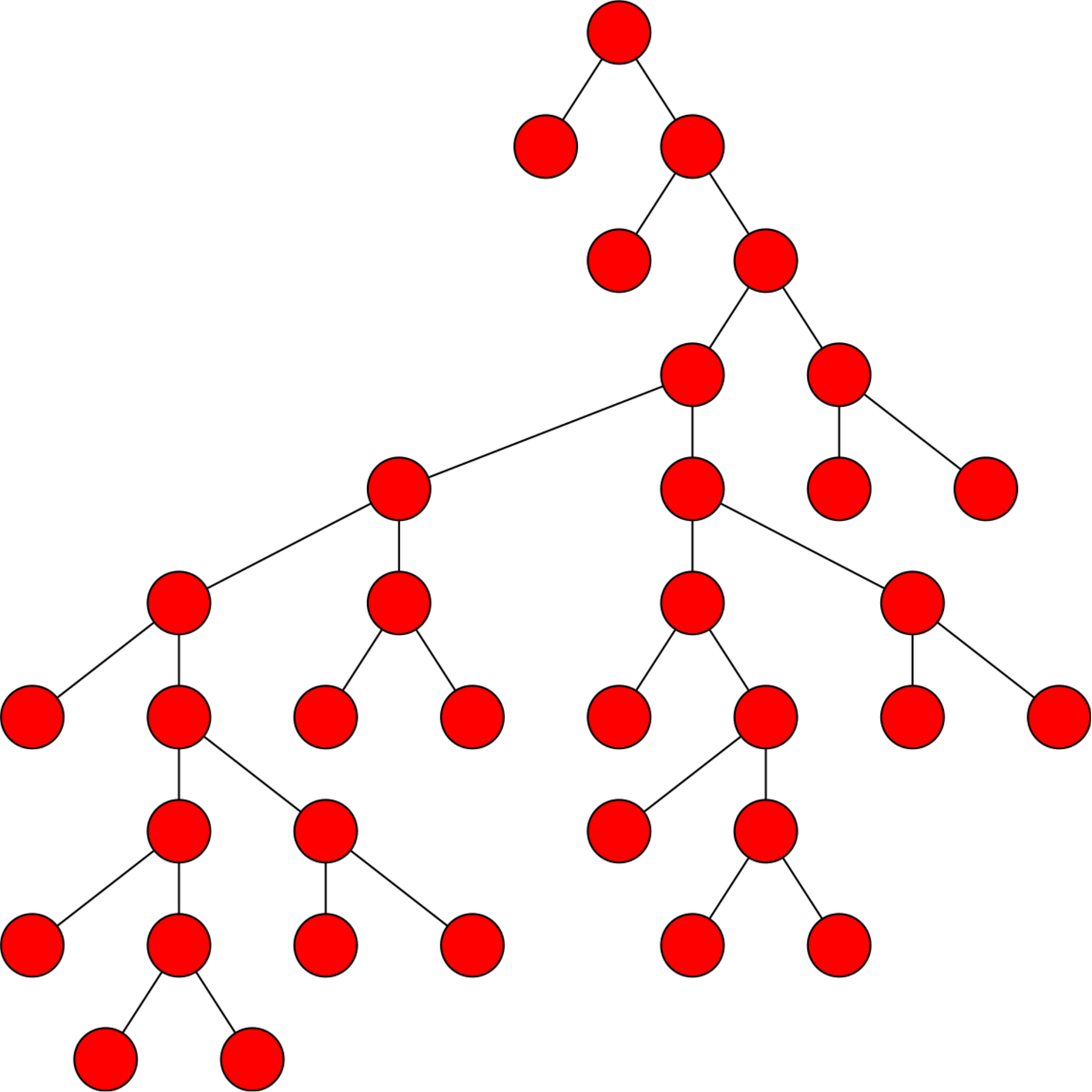}}
 }   	 
 \vskip0.1cm
 \caption{A random forest consisting of three trees.
 }
\label{fig:forest_example}
\end{figure}

A tree is built recursively starting from the root and a subset $T'\subseteq T$
of the training data.
Each node splits the available data into two subsets, which are used
to build two subtrees becoming the children of the node. 
A node becomes a leaf when the associated training data subset is
\emph{pure} (\ie, only instances with the same label are left) or some
other stopping criterion is fulfilled (\eg, a maximum tree depth is
reached). A simple example is given in
Figure~\ref{fig:forest_example}.
For an internal node corresponding to a subset $S \subseteq \trainset'$
of training instances, one searches for a splitting dimension $i$ and
a threshold $\theta$ that maximize the \emph{information gain} 
\begin{equation}
 G_{i,\theta}(S) = Q(S) - \frac{|L_{i,\theta}|}{|S|} Q(L_{i,\theta}) -\frac{|R_{i,\theta}|}{|S|} Q(R_{i,\theta})
\end{equation}
with $L_{i,\theta}=\{ (\vec x, y)\in\ S \,|\,   x_i \leq \theta \}$,
$R_{i,\theta}=\{ (\vec x, y)\in\ S \,|\,   x_i > \theta \}$, 
by minimizing the ``impurity''  of the subsets assessed by an \emph{impurity measure} $Q$~\cite{HastieTF2009,Murphy2012,Breiman2001}. 

\begin{algorithm}[t]
\caption{\textsc{BuildRandomForest($T$, $B$, $F$)}}
\label{alg:forest}
\small
\begin{algorithmic}[1]
\REQUIRE $T=\{(\vec{x}_{1},y_{1}), \ldots, (\vec{x}_{\tsize}, y_{\tsize})\} \subset \Reals^\tdim \times \Yspace$, $B \in \Nats$, and $F\in\{1,\ldots,\tdim\}$.
\ENSURE Trees $\mathcal{T}_1,\ldots,\mathcal{T}_B$ for $T$.
\FOR{$b=1,\ldots,B$}
\STATE Draw bootstrap sample $T'$ from $T$
\STATE $\mathcal{T}_b$ = \textsc{BuildTree}($T'$, F)
\ENDFOR
\STATE \textbf{return} $\mathcal{T}_1, \ldots, \mathcal{T}_B$
\end{algorithmic}%
\end{algorithm}%

\begin{algorithm}[t]
\caption{\textsc{BuildTree}($S$, $F$)}
\label{alg:cart}
\small
\begin{algorithmic}[1]
\REQUIRE Set $S \subseteq \trainset$ and $F\in\{1,\ldots,\tdim\}$.
\ENSURE Tree $\mathcal{T}$ built for $S$
\IF{$S$ is pure (or some other criterion fulfilled)}
\STATE \textbf{return} leaf node
\ENDIF
\STATE $(i^*,\theta^*) = \argmax_{i\in \{i_1, \ldots, i_f\} \subseteq \{1,\ldots,d\}, \theta} G_{i,\theta}(S)$
\STATE $\mathcal{T}_l$ = \textsc{BuildTree}($L_{i,\theta}$)
\STATE $\mathcal{T}_r$ = \textsc{BuildTree}($R_{i,\theta}$)
\STATE Generate node storing the pair $(i^*,\theta^*)$ and pointers to its subtrees $\mathcal{T}_l$ and $\mathcal{T}_r$. Let $\mathcal{T}$ denote the resulting tree.
\STATE \textbf{return} $\mathcal{T}$
\end{algorithmic}%
\end{algorithm}%

The overall construction of a random forest is sketched in
Algorithm~\ref{alg:forest}.
An ensemble model exploits the diversity of its members.
For standard random forests, one usually considers $\forestNTrees$
subsets of the training patterns. These \emph{bootstrap samples}
are drawn uniformly at random (with replacement) from $\trainset$ to
obtain slightly different training datasets and, hence,
trees. Another strategy to introduce randomness is to vary the
splitting mechanism at the internal nodes during the construction by,
\eg, considering different subsets of features for each node split
among which the one with the best splitting quality is selected (or by
considering random splitting thresholds, see below). The leaves of a
single tree store the label information. For regression problems, one
usually computes the mean of all labels associated with the leaf,
whereas the most frequent label is stored for classification scenarios
(or the distribution over the labels). The prediction $f_\forestTreeIndex (\vec{x})$ for a single tree is obtained by traversing the tree from top to bottom based on the splitting thresholds stored in the internal nodes until a leaf node is reached. The associated label then determines the prediction of the tree.

\subsection{Large-Scale Construction}
The construction discussed so far corresponds to standard random forests as proposed by Breiman~\cite{Breiman2001}. Various alternatives have been suggested over the past years. A popular one is the concept of \emph{extremely randomized trees}~\cite{GeurtsEW2006}, which is based on ``random'' thresholds for each feature $i$ in Line~$4$ of Algorithm~\ref{alg:cart} (more precisely, a random value between the minimum and the maximum is considered). In practice, resorting to these potentially ``suboptimal'' splits often yields competitive and sometimes even superior tree ensembles. Further, training such variants might be faster.


Many different random forest implementations have been proposed during the past years. 
The efficiency of the recursive construction of the individual trees via \textsc{BuildTree} heavily depends on the particular random forest variant that is considered and on the heuristics being used to speed up the process.
Today's state-of-the-art software makes use of various other
implementation tricks to reduce the practical
runtime.\footnote{For instance, one can keep track of ``locally
  constant'' features (\ie, a dimension $i$ does not need to be
  checked anymore for $S$ and all the descendant nodes in case all
  patterns in $S$ exhibit the same value w.r.t.~dimension $i$). Further, it is beneficial to represent a bootstrap sample $\trainset'$ via weights instead of
    duplicating training instances
~\cite{Louppe2014}. A popular well-engineered implementation based on highly-tuned C code is provided by the \emph{Scikit-Learn}~\cite{scikit-learn} package.}
A recent trend in data mining is to make use of massively-parallel
devices such as graphics processing units (GPUs) to accelerate the
generation and application of random
forests~\cite{EssenMGP2012,GrahnLLS2011,JanssonSB2014,Sharp2008}. However,
aiming at full, unpruned trees, such approaches do not seem to improve over a standard multi-core execution. Another line of research considers variants that are tailored towards special cases. For instance, Louppe and Geurts~\cite{LouppeG2012} consider small subsets of the data, called \emph{patches}. Each patch is based on a different subset of features and the overall ensemble consists of trees built independently on the patches. 

In addition, distributed construction schemes have been proposed that
are based on, \eg, \emph{MapReduce}~\cite{DeanG2008}. Several
strategies to implement random forests via the MapReduce framework are
described by del R\'{i}o~\etal~\cite{delRioLBH2014}. A natural one
(which is also implemented by the {Apache Mahout}\textsuperscript{TM}
library) is to consider subsets of the data and to build individual
trees/forests for each of these subsets; the overall ensemble is then
composed of all individual trees. The PLANAT
implementation~\cite{PandaHBB2009} also resorts to MapReduce. Similarly to our work, it also stops the recursive construction as soon as the subsets become small enough to be handled by a single machine. However, the overall implementation aims at distributed computing environments and the construction of the upper parts of the trees are handled in a conceptually very different way. Further, the trees are built independently from each other.


Finally, efficient implementations exist for the related task of computing ensembles of boosted trees~\cite{ChenG2016}. These ensembles, however, rely on many shallow trees that are built in an iterative fashion. Accordingly, such implementations do no not necessarily perform well when building deep, fully-grown trees. 

\subsection{Deep Trees}
\label{sec:deep_trees}
The original random forest implementation proposed by Breiman~\cite{Breiman2001} grows full trees. 
A simple variant is to build trees up to a certain depth only. While the validation and test accuracies might still be good, the induced forest might loose its capability to deal with rare classes.
The so-called $m$-out-of-$n$ subset strategy partitions the dataset into subsets and builds individual trees/forests for these subsets. While one makes use of all the data, this strategy might yield non-optimal results as well. As discussed by Genuer~\etal~\cite{GenuerPTV2015} and del R\'{i}o~\etal~\cite{delRioLBH2014}, such subset strategies usually cause a shift towards the dominant classes and instances (in a certain region of the feature space). For example, assume that one is given a single ``rare'' instance (\ie, $P(y=c)$ very small for a class $c$) and assume that one considers a partition of the data into three equal-sized subsets. The rare object would only be  contained in one of the subsets and, hence, would never be predicted via the overall ensemble.

\begin{figure}[t]
\subfigure[Data]{
\resizebox{0.22\columnwidth}{!}{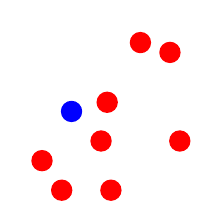}
}
\,
\subfigure[Partition (three subsets)]{
\resizebox{0.22\columnwidth}{!}{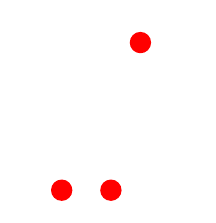}
\resizebox{0.22\columnwidth}{!}{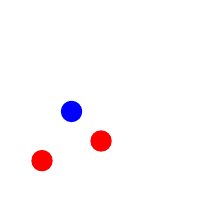}
\resizebox{0.22\columnwidth}{!}{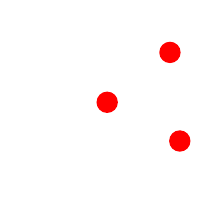}
}
\caption{
An ensemble built via the three subsets cannot identify the blue object anymore since two of the three models do not contain this instance. Such an effect can also be observed in case $\mathbf{P(y|\vec{x})}$ is unbalanced for a subregion~$\vec{x}$. 
}
\label{fig:sampling_motivation}%
\end{figure}%
Reweighting strategies applied in a post-processing phase aim at reducing these negative side-effects, see, \eg, del R\'{i}o~\etal~\cite{delRioLBH2014} for several adaptions. However, finding appropriate weights is a challenging task as well and such methods generally focus on shifts in $P(y)$ introduced by subsampling.
Such a bias towards a dominant class/label can also occur in a \emph{region of the feature space}, as sketched in Figure~\ref{fig:sampling_motivation}. Hence, even in case a reasonable amount of labeled instances is given for a rare class, subsampling strategies might lead the model to completely ignore this class, see 
Genuer~\etal~\cite{GenuerPTV2015} for a detailed discussion. 

%


\section{Algorithmic Framework}
\label{section:algorithmic_framework}
We propose a wrapper-based approach that can handle massive datasets given single compute node resources.


\subsection{Wrapper-Based Construction}
We start by outlining the construction of a single tree; the overall implementation simultaneously constructs  all trees, which is described in the next section. The basic idea is to build a ``top tree'' based on a small \emph{random} subset of the training data and to use this tree to obtain a partition of \emph{all} the training instances into (ideally) almost equal-sized subsets. 

The overall workflow is shown in Algorithm~\ref{alg:workflow}: In the first phase, a top tree $\tree$ is built for a small subset $\topsubset$ of $\trainset$ drawn uniformly at random (without replacement). Afterwards, all available training instances are distributed to the leaves of the top tree. That is, one determines for each training instance $(\vec{x}_i,y_i)$ the index of the leaf within the top tree the pattern $\vec{x}_i$ would be assigned to. Finally, in the third phase, one computes for each of the induced subsets $T_1, \ldots, T_M$ an associated bottom tree, which is then attached to the corresponding leaf of the top tree. This yields the final tree~$\tree$. Thus, the overall tree $\tree$ basically corresponds to a standard tree built via \textsc{BuildTree} with potentially slightly different splits conducted due to the random subset used for the top tree. A direct implementation of this approach, however, does not necessarily yield an efficient implementation due to, \eg, potentially very unbalanced top trees. The modifications needed to render this approach efficient are described next.

\begin{algorithm}[t]
\caption{\textsc{BuildBigTree}}
\label{alg:workflow}
\small
\begin{algorithmic}[1]
\REQUIRE $\trainset=\{(\vec{x}_1,y_1), \ldots, (\vec{x}_\tsize, y_\tsize)\} \subset \Reals^\tdim \times \Yspace$, $F \in \{1,\ldots,\tdim\}$, subset size $\topsubsetsize$, and leaf bucket size $\leafbucketsize$.
\ENSURE Tree $\tree$ built for $\trainset$
\STATE Retrieve random subset $\topsubset \subset \trainset$ with $|\topsubset|=\topsubsetsize$
\STATE $\tree_{top}$ = \textsc{BuildTopTree}$(\topsubset, \leafbucketsize)$
\STATE $\trainset_1, \ldots, \trainset_\numtoptreeleaves=\textsc{Distribute}(\tree_{top}, \trainset)$
\STATE $\tree = \emptyset$
\FOR{$j=1,\ldots,\numtoptreeleaves$}
\STATE $\tree = \tree \cup \textsc{BuildTree}(\trainset_j)$
\ENDFOR
\STATE \textbf{return} $\tree$
\end{algorithmic}%
\end{algorithm}%


\subsubsection{Construction of Top Trees}
\begin{algorithm}[t]
\caption{\textsc{BuildTopTree}}
\label{alg:toptree}
\small
\begin{algorithmic}[1]
\REQUIRE Set $\trainset'=\{(\vec{x}_{i_1},y_{i_1}), \ldots, (\vec{x}_{i_\topsubsetsize}, y_{i_\topsubsetsize})\} \subset \Reals^\tdim \times \Yspace$ leaf desired bucket size $\leafbucketsize$.
\ENSURE Root node of a binary tree $\tree_{top}$.
\STATE Create empty stack $\mathcal{P}$; create root node $\mathcal{N}_0$
\STATE $i_{\text{node}} = -1$
\STATE $\mathcal{P}.push((\trainset', \mathcal{N}_0))$
\WHILE{$\mathcal{P}$ is not empty}
\STATE $i_{\text{node}} = i_{\text{node}} +1$
\STATE $(\bar{\trainset}, \bar{\mathcal{N}})$ = $\mathcal{P}.pop()$
\IF{$|\bar{\trainset}| < \max(2, \leafbucketsize \cdot \frac{\topsubsetsize}{\tsize})$}
\STATE $\bar{\mathcal{N}}.value = i_{\text{node}}$
\STATE \textbf{return} $\bar{\mathcal{N}}$
\ENDIF
\STATE
$ (j^{*},\theta^{*}) = \argmax_{(j,\theta)} \bar{G}_{i,\theta}(S)$
\STATE Split $\trainset'$ into ${\trainset'_l}$ and ${\trainset'_r}$ according to $(j^{*},\theta^{*})$
\STATE Create left node $\bar{\mathcal{N}_l}$ of $\bar{\mathcal{N}}$ and $\mathcal{P}.push((\trainset'_l, \mathcal{N}_l))$
\STATE Create right node $\bar{\mathcal{N}_r}$ of $\bar{\mathcal{N}}$ and $\mathcal{P}.push((\trainset'_r, \mathcal{N}_r))$
\ENDWHILE

\STATE \textbf{return} $\mathcal{N}_0$
\end{algorithmic}%
\end{algorithm}%

Since the top tree is only built on a small subset, different and potentially ``suboptimal'' splits might be considered compared to a direct construction via \textsc{BuildTree}. Note, however, that the notion of ``optimal'' is anyway vague in this context. For instance, extremely randomized trees, which resort to simple random splitting thresholds, often even yield competitive or even superior overall ensembles compared to their counterparts that rely on ``optimal'' splitting thresholds.

In practice, simply resorting a standard $\textsc{BuildTree}$ construction scheme might not yield a feasible approach. This is due to the fact that, for some datasets, the top trees might become very unbalanced, leading to many ``small'' leaves that do not contain many training instances after the distribution phase. In addition, the standard splitting scheme might yield very big leaves after the distribution phase; while these leaves might be ``pure'' after the construction phase of the top tree, they might become unpure again after the distribution phase. This actually is a problem since the induced big leaves still have to be processed in the third phase---and this might not be possible given the restricted resources (e.g., such a big leaf might contain hundreds of millions of patterns).

For this reason, we adapt the construction of the top tree, see Algorithm~\ref{alg:toptree}: The workflow is essentially the same except for the following two minor yet crucial modifications: 
\begin{enumerate}
 \item \emph{Minimal leaf size stopping:} Firstly, the recursive construction \emph{only} stops as soon as the minimal leaf size is reached. By doing so, we ensure that the resulting leaf buckets are small enough for the further processing (otherwise, almost pure leaves might yield very big leaf buckets). The parameter $\leafbucketsize$ specifies the desired maximum size of a leaf bucket \emph{after} the distribution phase. Since the actual number of assigned instances is only known after the distribution of all instances, we resort to an estimate $\bar{\leafbucketsize} = \max(2, \leafbucketsize \cdot \frac{\topsubsetsize}{\tsize})$ for $\leafbucketsize$. 
 \item \emph{Balanced splits:} Secondly, to handle degenerated cases, we consider the following modified information gain criterion $\bar{G}_{i,\theta}(S)$ that favors balanced partitions in the top tree:
 \begin{equation}
 \label{eq:adapted_gain}
 \bar{G}_{j,\theta}(S) = (1-\lambda) G_{j,\theta}(S) - \lambda \frac{\left | |L_{j,\theta}|-|R_{j,\theta}| \right |}{|S|}
\end{equation}
The second part in the above objective favors ``balanced'' partitions, which are similar to those done for standard $k$-d trees~\cite{Bentley1975}.\footnote{Note that using the median does not necessarily yield almost equal-sized partitions.} The parameter $\lambda \in [0,1]$ determines the tradeoff between the standard information gain and favoring balanced partitions. 
\end{enumerate}
Finally, no bootstrap samples are drawn for the construction of top trees as well as optimal splits w.r.t. (\ref{eq:adapted_gain}) are considered. Note that the adapted information gain is especially important for splits of almost pure nodes. Here, the first part of (\ref{eq:adapted_gain}) will yield similar gains for various thresholds. However, the second part will enforce the splits to be balanced.

\begin{figure*}[t]
\centering
\subfigure[]{\resizebox{0.9\textwidth}{!}{\includegraphics{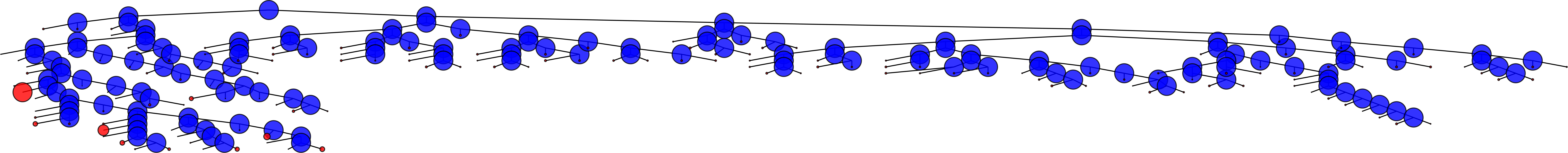}}}
\subfigure[]{\resizebox{0.9\textwidth}{!}{\includegraphics{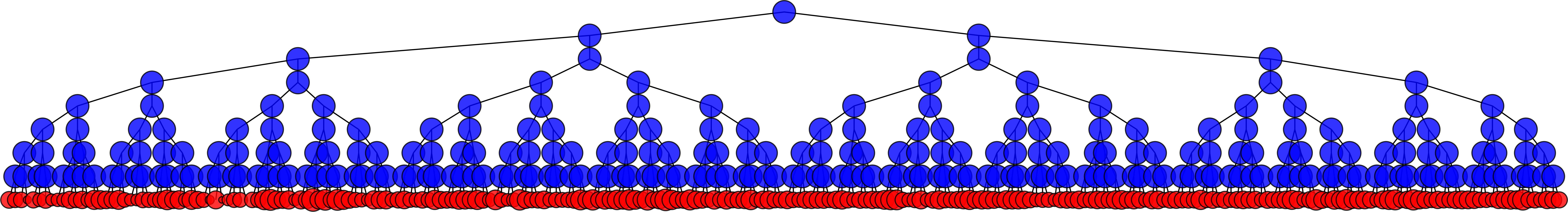}}}
\vspace{-2ex}
\caption{Two top trees built via \textsc{BuildTopTree} using $\mathbf{\bar{G}_{j,\boldsymbol{\theta}}(S)}$ with (a) $\mathbf{\boldsymbol{\lambda}=0}$ and (b) $\mathbf{\boldsymbol{\lambda}=1}$, respectively. The size of a leaf (red) is proportional to the number of points assigned to it. The standard construction scheme that stops as soon as a node is pure and which resorts to the normal information gain ($\mathbf{\boldsymbol{\lambda}=0}$) might yield very unbalanced leaves (a few leaves contain almost all instances!). The adapted construction scheme with $\mathbf{\boldsymbol{\lambda}=1}$ yields very balanced partitions. In expectation, all these leaves will contain about $\mathbf{M}$ leaves after the distribution phase and are, hence, small enough for the construction of bottom trees.}
\label{fig:top_trees}
\end{figure*}
An example for a very unbalanced tree with a single leaf containing most of the patterns is given in Figure~\ref{fig:top_trees} (which is based on the \landsat dataset, see Appendix~\ref{appendix:landsat}). The few very big leaves depict a problem since they might become unpure again after the distribution phase and, hence, still have to be considered and processed in the third phase. The adapted construction scheme outlined above with $\lambda=1$ yields almost equal-sized partitions, all being sufficiently ``small'' such that the leaves will contain about $M$ patterns \emph{after} the distribution phase. 

Note that the adapted splitting scheme actually continues splitting up pure nodes until a leaf size of $\bar{M}$ is reached. While this seems like an unnecessary operation, it is crucial to obtain manageable partitions for the third phase, the construction of the bottom trees. A potential drawback of this approach is that more splits than needed are actually conducted. Since one only knows about the properties of the final leaves after the distribution phase, this cannot be avoided. Further, favoring balanced partitions using, e.g., $\lambda=1$, might yield to ``similar'' top trees and, hence, less randomness in the upper parts of the final trees. Also, the splits conducted might suboptimal w.r.t. the original information gain criterion $G_{j,\theta}(S)$, which might yield to suboptimal top parts of the trees (e.g., no feature selection is conducted).\footnote{For large-scale scenarios with millions or even billions of training patterns, these potential drawbacks do not seem to have a significant influence (basically, only a few suboptimal splits are conducted in the upper parts of the generally very deep trees). Note that the adapted node splitting (in case $\lambda=1$ is used) is related to Mondarian Forests~\cite{LakshminarayananRT2014}, which only resort to label-independent node splits.}

\subsubsection{Distribution \& Construction of Bottom Trees}
\begin{algorithm}[t]
\caption{\textsc{Distribute}}
\label{alg:distribute}
\small
\begin{algorithmic}[1]
\REQUIRE A set $\trainset=\{(\vec{x}_{1},y_{1}), \ldots, (\vec{x}_{\tsize}, y_{\tsize})\} \subset \Reals^\tdim \times \Yspace$ of training patterns and a top tree $\tree_{top}$.
\ENSURE A partition $\trainset_1, \ldots, \trainset_\numtoptreeleaves$ of $\trainset$.
\STATE $LI=\textsc{GetLeavesIndices}(\trainset)$
\STATE $\trainset_1, \ldots, \trainset_\numtoptreeleaves=\textsc{Partition}(LI, \trainset)$
\STATE \textbf{return} $\trainset_1, \ldots, \trainset_\numtoptreeleaves$
\end{algorithmic}%
\end{algorithm}%
Given the top tree, all training instances are distributed to the top tree's leaves, see Algorithm~\ref{alg:distribute}. The top tree is modified in such a way that the leaf index is returned for a query instead of a (label-based) prediction. The resulting indices can then be used to partition all the training data $\trainset$ to the different leaf buckets. 

Finally, one or more bottom trees are built for each of the leaf buckets $\trainset_1, \ldots, \trainset_\numtoptreeleaves$. The number $\nbottomtrees \geq 1$ of bottom trees built per bucket can be defined by the user. For large-scale scenarios, sharing top trees among the different overall trees can be computationally very advantageous since less calls to \textsc{Distribute} are needed (hence, passes over the data) and since the construction of all bottom trees can effectively be parallelized (using the same chunk of data fitting in the system's memory). However, sharing top trees that way generally leads to less randomness in the overall ensemble, which might reduce the model's quality. From a practical perspective, this depicts a trade-off between runtime and tree diversity. 



\subsection{Implementation}
\label{subsec:implementation_details}
The wrapper-based construction outlined above yields much smaller partitions associated with the leaves of the top tree, which can be \mbox{handled} more efficiently. Its efficiency, however, depends on a careful implementation of the involved steps.

Some of the steps are conducted simultaneously for the different trees to be built. More precisely, the construction of the top trees as well as the distribution of the patterns are done via two single passes over the training instances. Each pass is conducted by processing all the data in chunks using a certain chunk size $\chunksize$ (\eg, $\chunksize=1,000,000$). In the first pass over the data, random subsets are extracted for the top trees. Given the subsets, the top trees are built, which are then used to distribute all instances to the top trees' leaves. Note that considering random subsets (based on a full pass over the data) might be crucial for obtaining good estimates $\bar{\leafbucketsize} = \max(2, \leafbucketsize \cdot \frac{\topsubsetsize}{\tsize})$. During the distribution phase, a new subset of training instances is created for each leaf bucket, which is either stored in memory or on disk (using HDF5~\cite{hdf5}). 

The construction of all top trees can be done using $\mathcal{O}(\topsubsetsize + \chunksize)$ memory, where $\topsubsetsize$ is the size of a single random subset and $\chunksize$ the chunk size. Further, the distribution of the instances can be done spending $\mathcal{O}(\topsubsetsize + \chunksize)$ additional memory as well. Finally, for the third phase, one needs $\mathcal{O}(\hat{\leafbucketsize})$ memory with $\hat{\leafbucketsize}$ being the maximal size of a leaf bucket after the distribution phase. 

The general wrapper-based framework is implemented in \texttt{Python} (version 2.7.11), where the \emph{Numpy} package (version 1.11.2)~\cite{scipy} is used for all matrix/vector-based operations. For the computation of both the top and the bottom trees, we resort to a pure \texttt{C} implementation that follows the construction scheme implemented by the \emph{Scikit-Learn} package (version 0.18.1)~\cite{scikit-learn}; \emph{Swig}~\cite{Beazley:1996:SEU:1267498.1267513} is used to generate a \texttt{Python} extension. The overall implementation is made publicly available on \url{https://github.com/gieseke/woody} under the \emph{GNU General Public License v3.0}.

\section{Experiments}
\label{section:experiments}
We consider a standard multi-core machine for all experiments and compare our approach with two state-of-the-art competitors. The experiments can be reproduced using the code made available on \url{https://github.com/gieseke/woody}.

\begin{figure*}[t!]
\centering
\subfigure[$M=1,000$]{
\resizebox{0.32\textwidth}{!}{\includegraphics{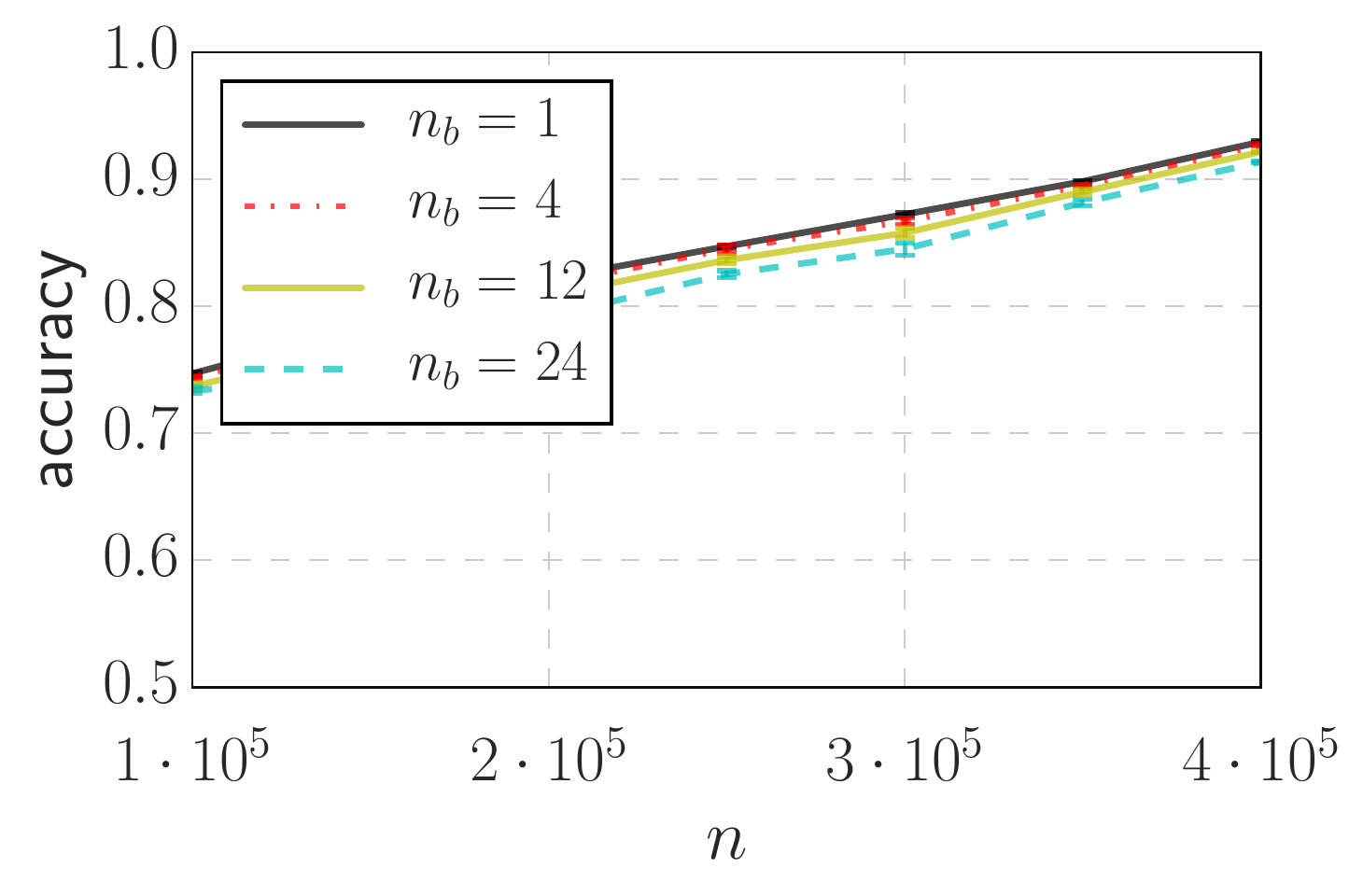}}
}
\subfigure[$M=10,000$]{
\resizebox{0.32\textwidth}{!}{\includegraphics{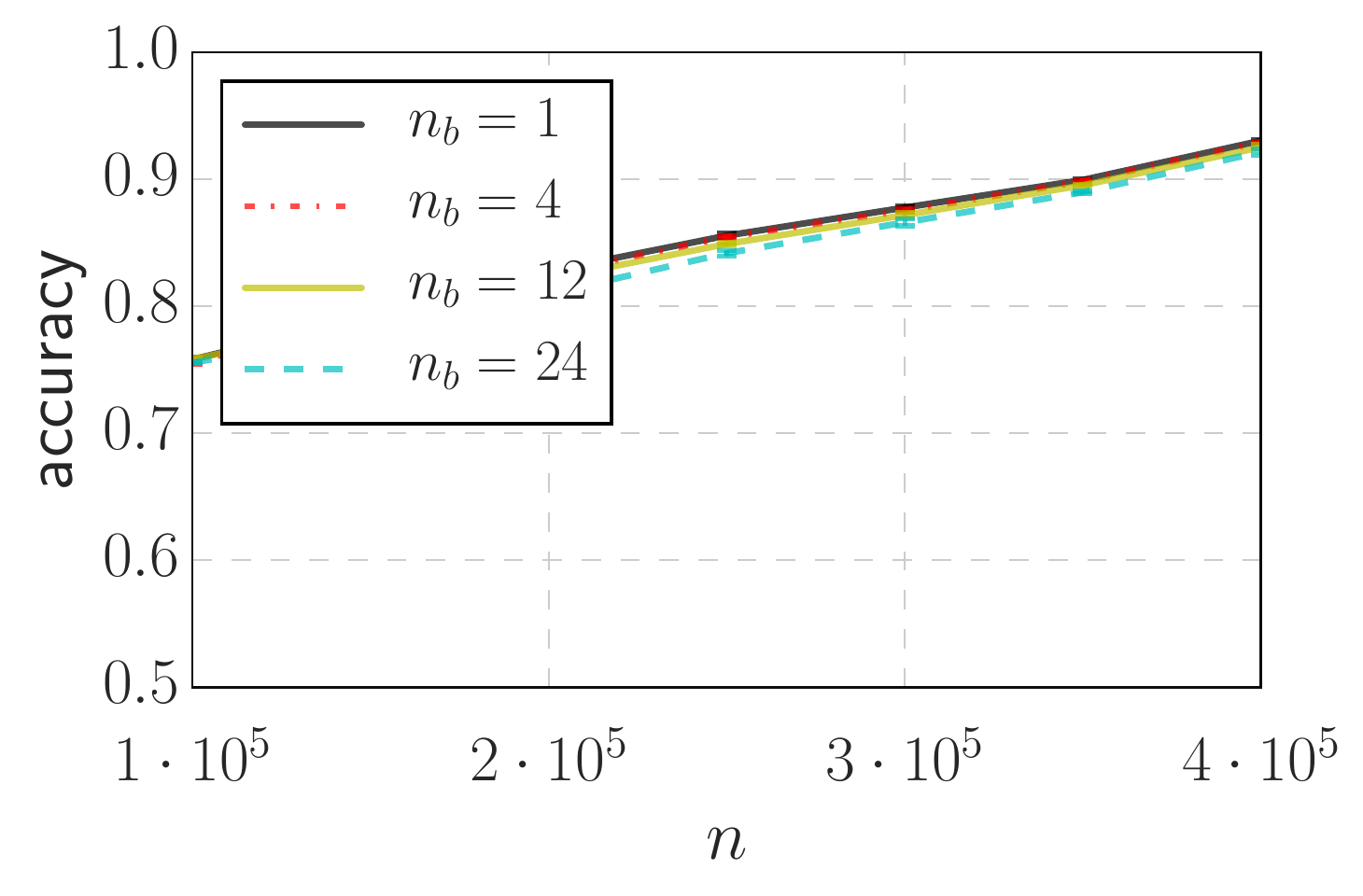}}
}
\subfigure[$M=75,000$]{
\resizebox{0.32\textwidth}{!}{\includegraphics{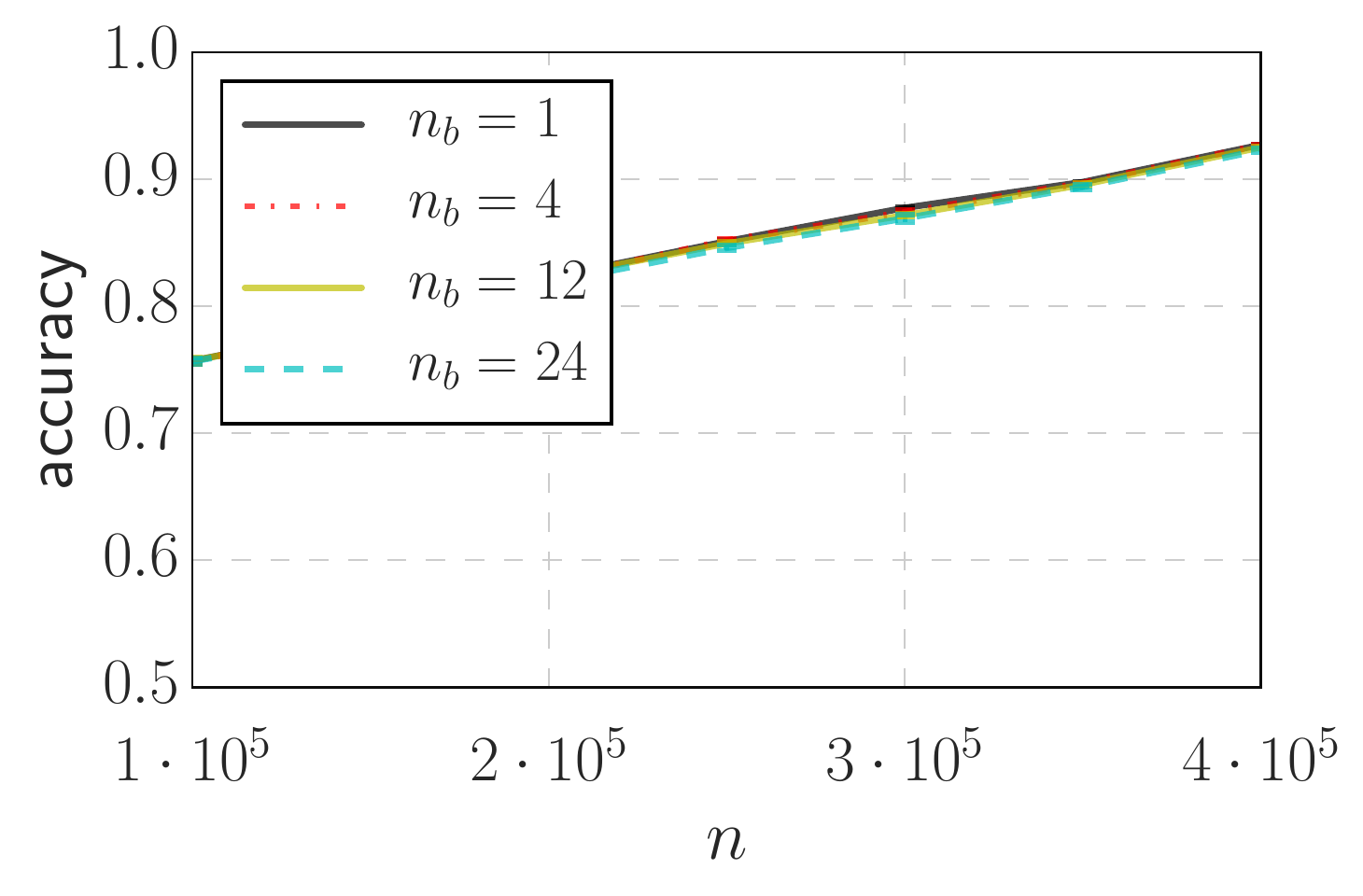}}
}
\vspace{-2ex}\caption{Influence of the number $\mathbf{\nbottomtrees}$ of bottom trees per top tree on the accuracy given the \covtype dataset. For each result (line), 24 trees are built in total. Three different \woody instances are considered that are induced by different assignments for~$\mathbf{M}$.
}
\label{fig:influence_n_bottom}
\end{figure*}

\newcommand{\tra}{\ensuremath{n_{\text{train}}}}
\newcommand{\tes}{\ensuremath{n_{\text{test}}}}
\newcommand{\da}{\ensuremath{d}}

\begin{table}[t]
\caption{Datasets}
\label{tab:data_sets}
\centering
\begin{tabular}{@{}lrrr@{}}
\toprule
Name & \tra & \tes & \da\\\midrule
\covtype & 464,809 & 116,203 &  54\\
\susy & 5,000,000 & 500,000 & 18\\
\higgs & 11,000,000 & 1,000,000 & 28\\
\landsat & 1,000,000,000 & 2,964,607 & 81\\
\bottomrule
\end{tabular}
\end{table}

\subsection{Experimental Setup}
All experiments were conducted on a standard desktop computer with an \texttt{Intel(R) Core(TM) i7-3770} CPU~at 3.40GHz (4 cores, 8 hardware threads), 16GB~RAM, and 16GB~swap space. The operating system was \texttt{Ubuntu 16.04} (64 Bit). In general, we focus on runtimes for the construction phases as well as on the accuracies obtained on the test set. If not stated otherwise, all results reported are averages over four runs with different seeds being used for the initialization. Our approach can, in principle, be applied to any kind of random forest variant (\eg, extremely randomized trees). For the sake of simplicity, we resort to standard random forests as sketched in Section~\ref{section:background}. For the experiments, we consider the following four implementations:

\begin{enumerate}
 \item The first one is the wrapper-based construction scheme proposed in this work, referred to as \woody.
 \item The second one, \subsetwood, uses subsets drawn uniformly at random from \emph{all} the available training instances. For each such subset, a single classification tree is built. The overall implementation resembles the wrapper-based implementation outlined above, \ie, random subsets are extracted via a pass over all instances. However, instead of top trees, standard trees are built for these subsets. The remaining points are \emph{not} distributed/considered. Intermediate results are stored on disk, as it is done for \woody.
 \item The third one, \sklearn, is the \texttt{RandomForestClassifier} implementation provided by the \emph{Scikit-Learn}~\cite{scikit-learn} package.
 \item The fourth one, \htwo, is the implementation provided by the \HTwoO package (\texttt{H2ORandomForestEstimator}).\footnote{\url{http://docs.h2o.ai}}
\end{enumerate}

Most parameters are set to their default values. Some of the parameters are automatically adapted to the specific dataset at hand, see Appendix~\ref{appendix:parameters} for the details. We focus on classification scenarios to
assess the classification performances. In all cases, we consider a
separate test set for evaluating the classification accuracy. For the
different experiments, \tra~training instances, \tes~test
instances, and \da~features are considered, see Table~\ref{tab:data_sets}. Despite the three medium-sized datasets \covtype, \susy, and \higgs~\cite{Lichman:2013}, we also consider \landsat, a large-scale dataset from the field of remote sensing containing up to one billion training instances, see Appendix~\ref{appendix:landsat} for details.


\subsection{Model Parameters}
We start by analyzing the influence of two of the main model parameters introduced by the wrapper-based scheme.

\subsubsection{Influence of $\nbottomtrees$}
The initial two phases can consume a significant part of the overall runtime. Especially the distribution phase involves extracting many large subsets that might have to be stored on disk. To reduce the overhead for these two phases, one can construct $\nbottomtrees>1$ bottom trees per top tree. This essentially leads to final trees sharing upper parts, which, in turn, might lead to less randomness in the overall ensemble. To investigate the influence of this parameter, we consider the \covtype~dataset and three different instances of \woody induced by different assignments for $M$: $M=1,000$, $M=20,000$, and $M=75,000$. If $M$ is small, larger top trees will be built. For large $M$, the top trees will have small sizes. Hence, we expect a slightly worse performance for small $M$ and a competitive performance for large $M$. 

We consider $1,4,12$ and $24$ as assignments for $\nbottomtrees$. In all cases, $24$ are built in total (\eg, $\nbottomtrees=24$ and $\ntoptrees=1$). 
The results are shown in Figure~\ref{fig:influence_n_bottom}. As expected, the accuracies are slightly worse for large $\nbottomtrees$. However, the differences are generally very small, indicating that sharing upper parts does not significantly hurt the performance. Further, the differences between the three instances of \woody for varying $M$ are very small as well, which indicates that sharing larger parts does not lead to a significant drop w.r.t. the accuracy as long as sufficiently large bottom trees are built.

\subsubsection{Influence of $\lambda$}
\begin{figure}[t!]
\centering
\subfigure[Runtimes]{
\resizebox{0.48\columnwidth}{!}{\includegraphics{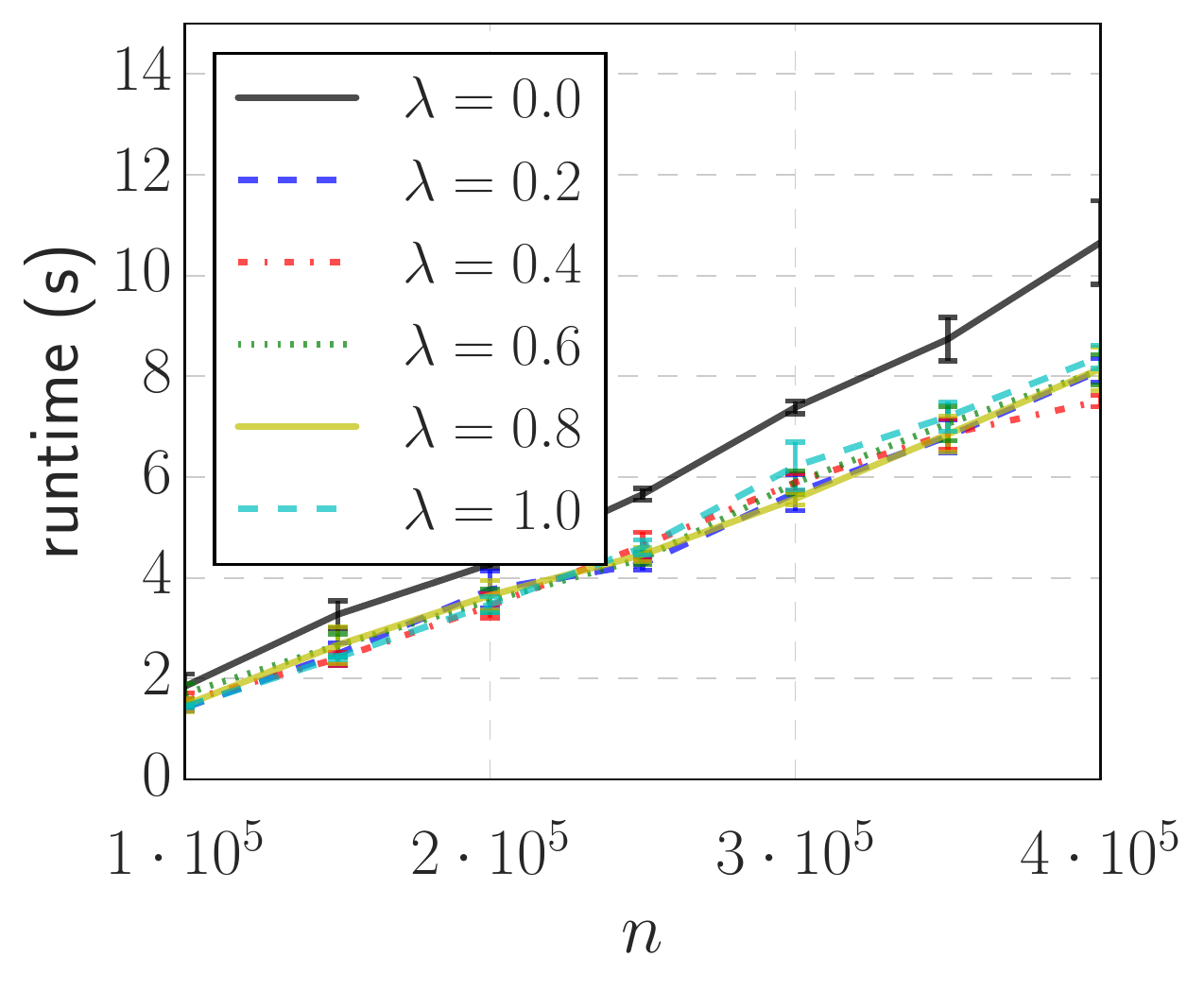}}
\resizebox{0.48\columnwidth}{!}{\includegraphics{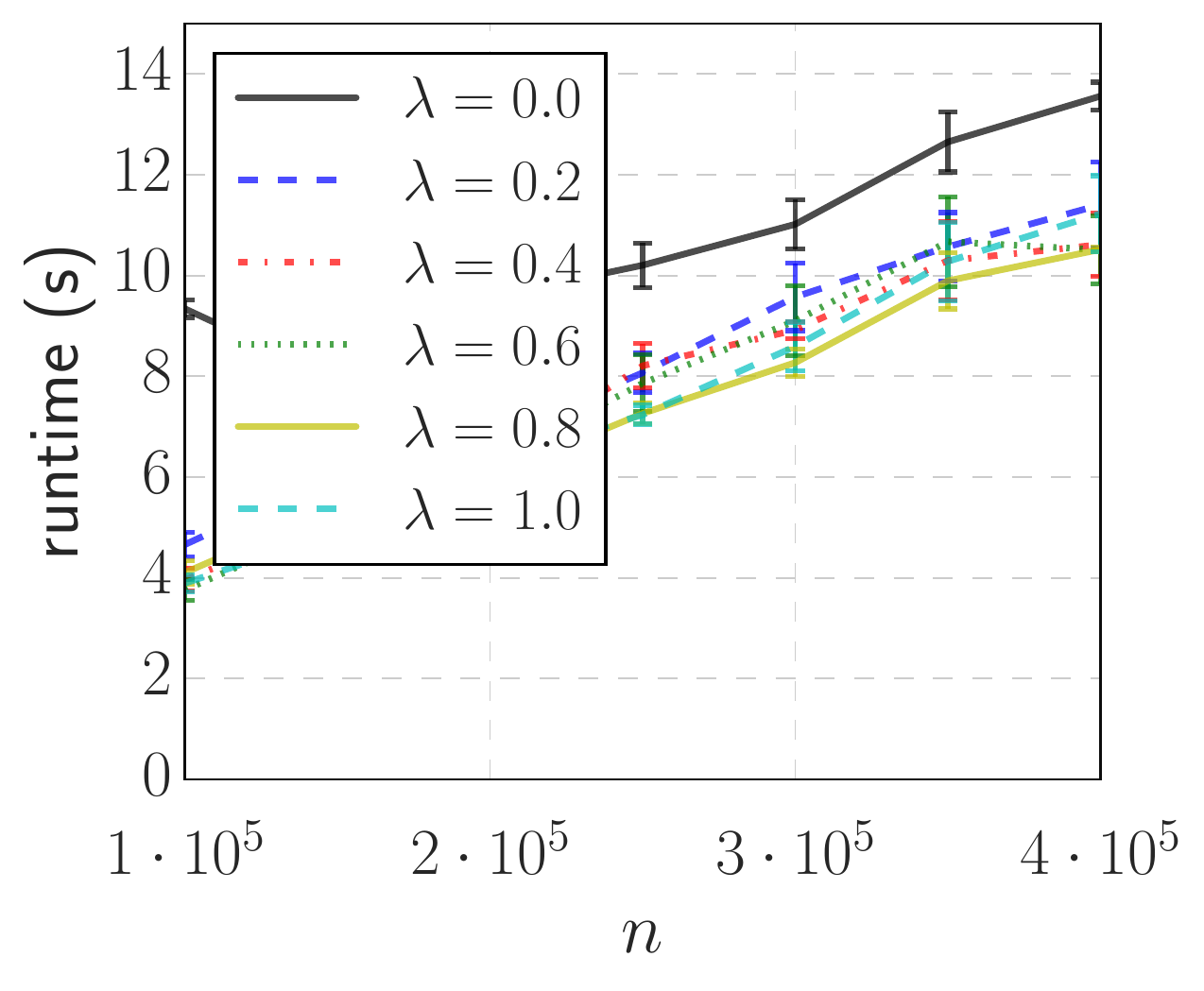}}
}
\subfigure[Accuracies]{
\resizebox{0.48\columnwidth}{!}{\includegraphics{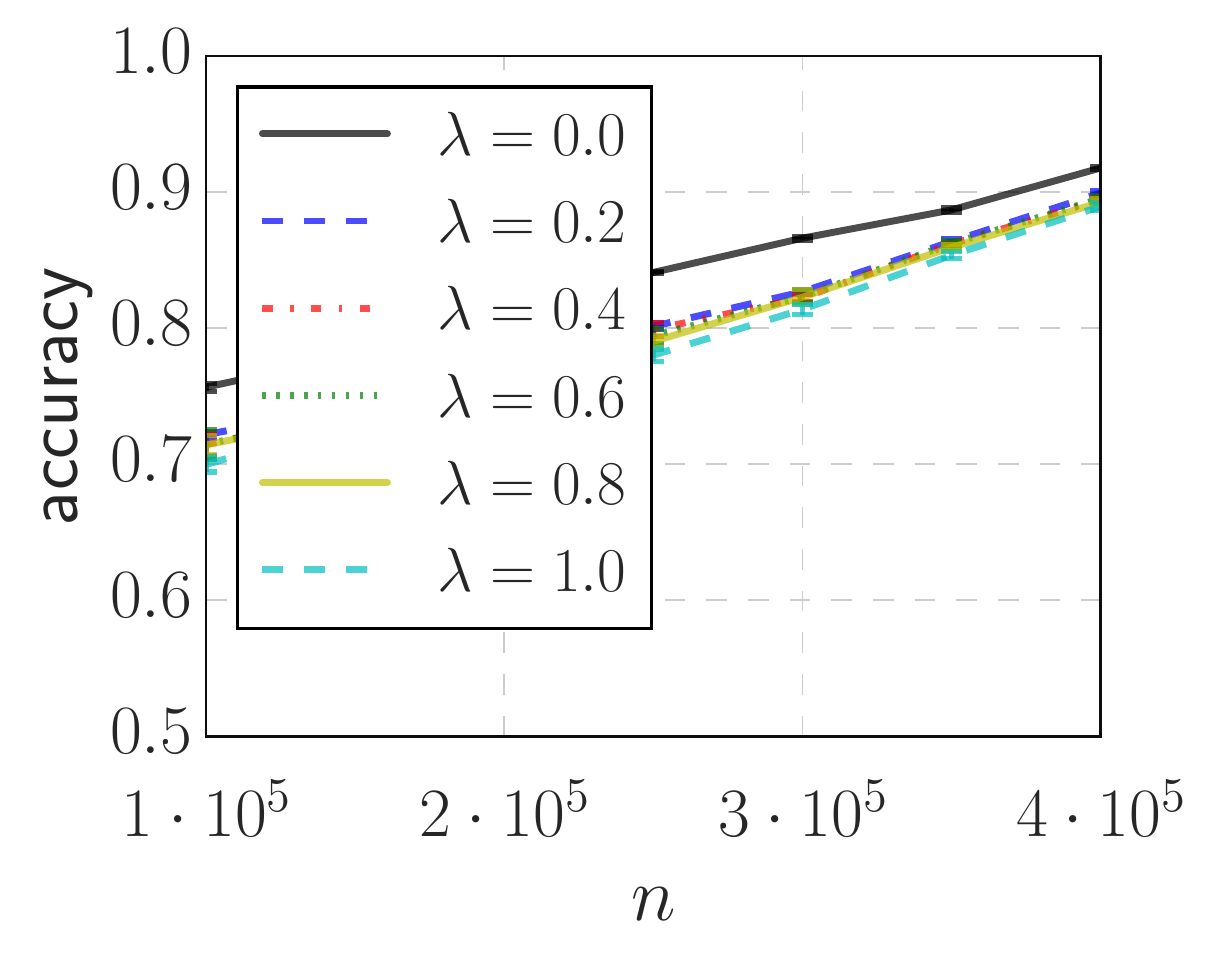}}
\resizebox{0.48\columnwidth}{!}{\includegraphics{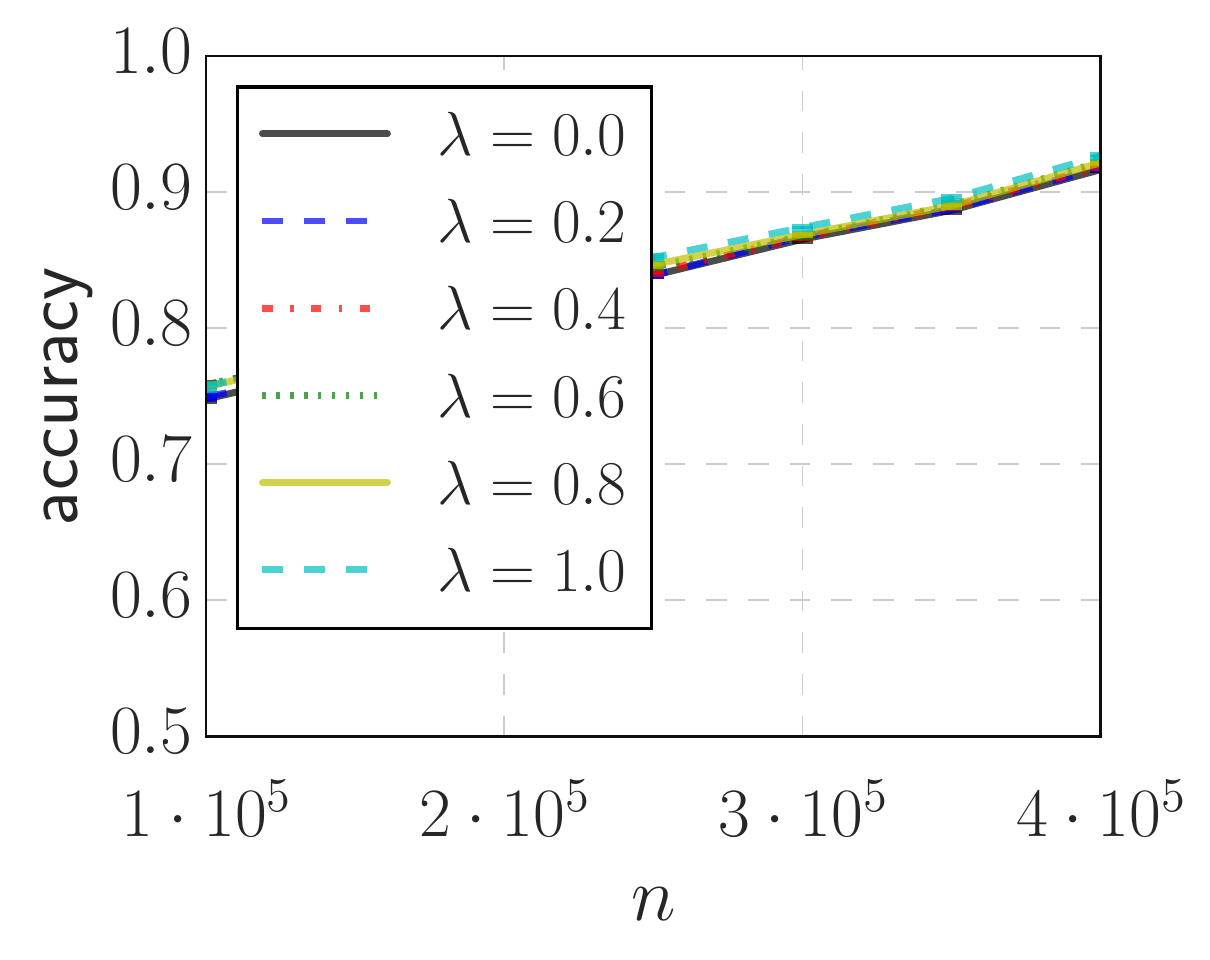}}
}
\vspace{-2ex}\caption{Influence of $\boldsymbol{\lambda}$. Left: Standard random forest with adapted information gain $\mathbf{\bar{G}}$. Right: The wrapper-based approach \woody with the top tree being built using $\mathbf{\bar{G}}$.}
\label{fig:influence_lambda}
\end{figure}

\begin{figure*}[t!]
\centering
\subfigure[\covtype]{
\resizebox{0.31\textwidth}{!}{\includegraphics{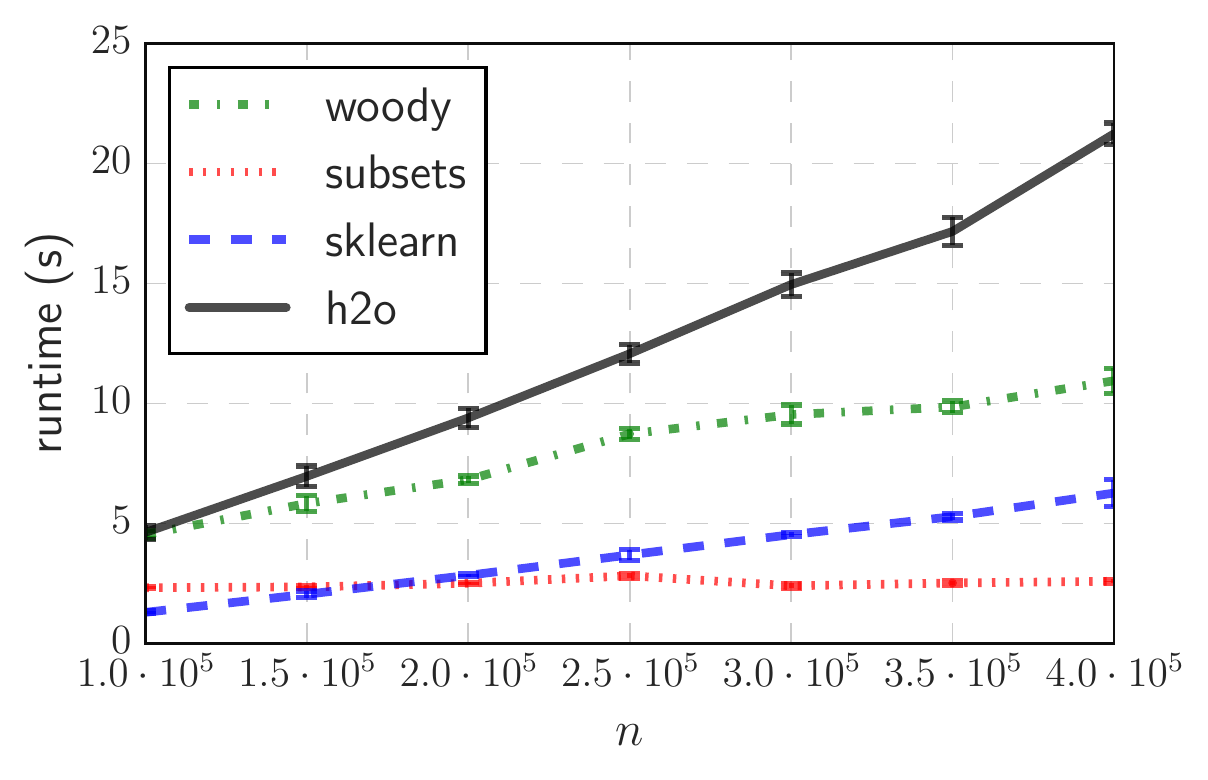}}
}
\subfigure[\susy]{
\resizebox{0.31\textwidth}{!}{\includegraphics{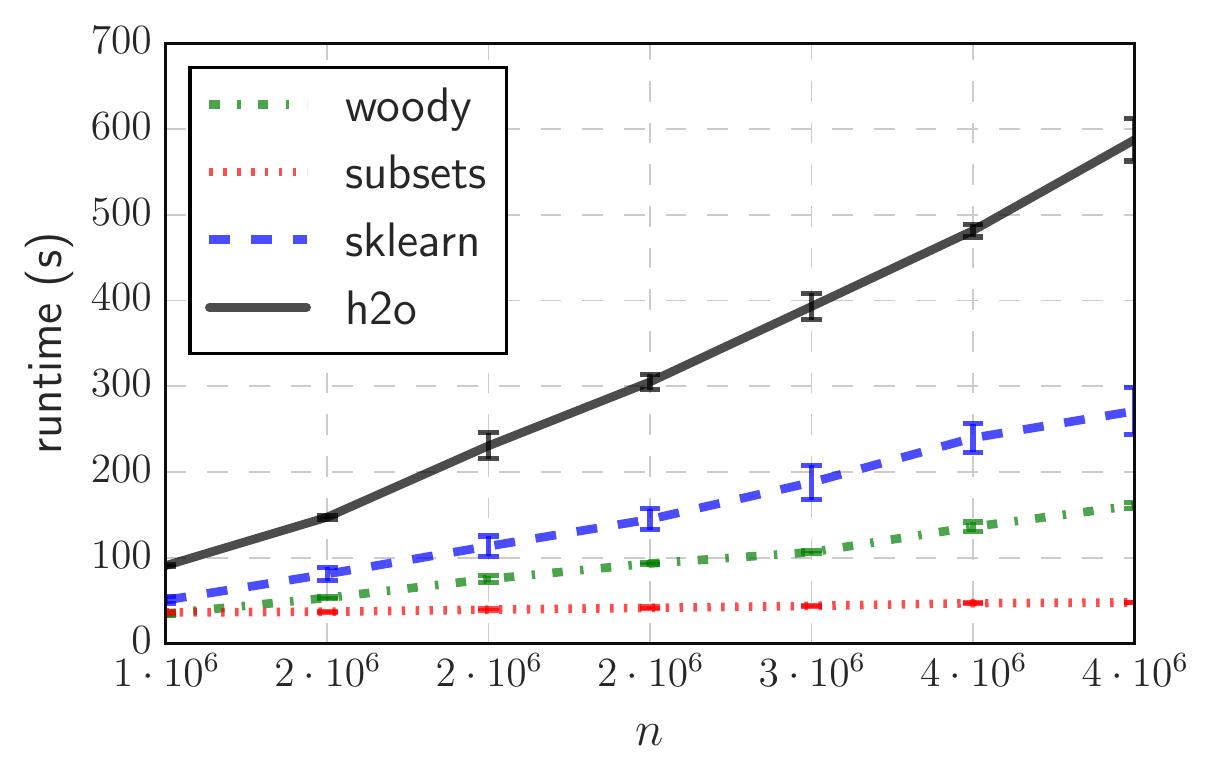}}
}
\subfigure[\higgs]{
\resizebox{0.31\textwidth}{!}{\includegraphics{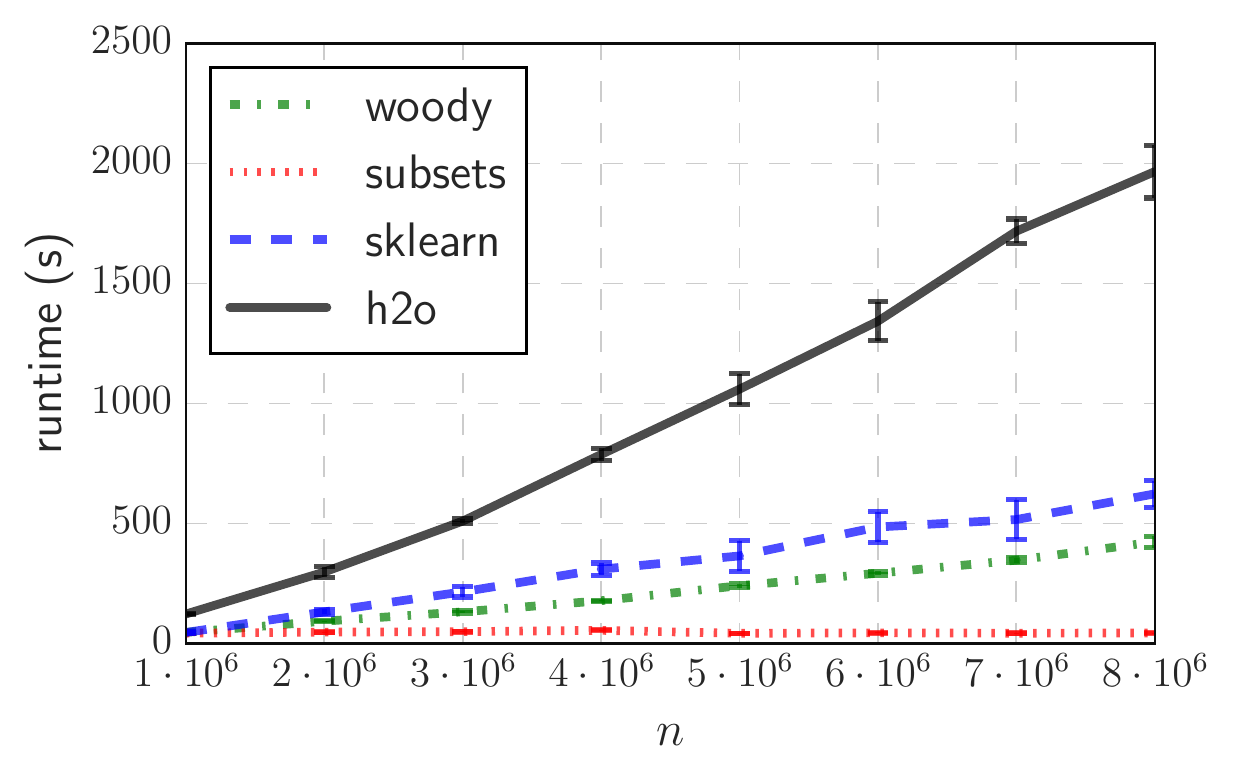}}
}
\subfigure[\covtype]{
\resizebox{0.31\textwidth}{!}{\includegraphics{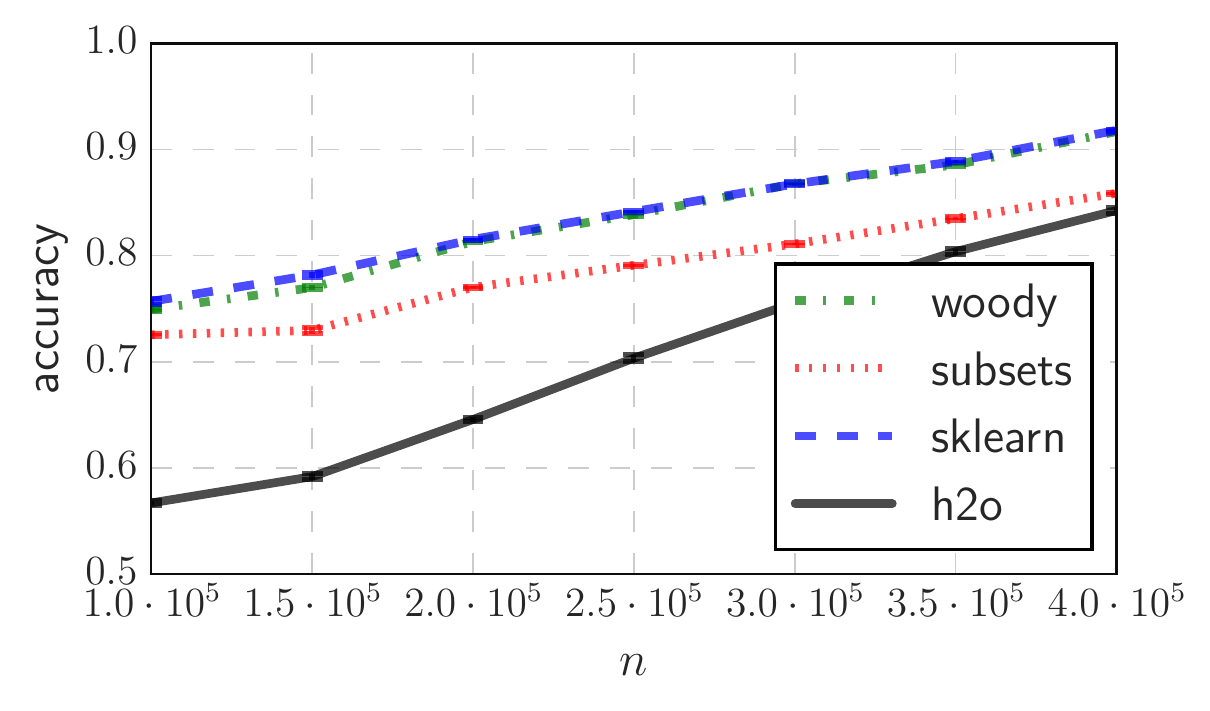}}
}
\subfigure[\susy]{
\resizebox{0.31\textwidth}{!}{\includegraphics{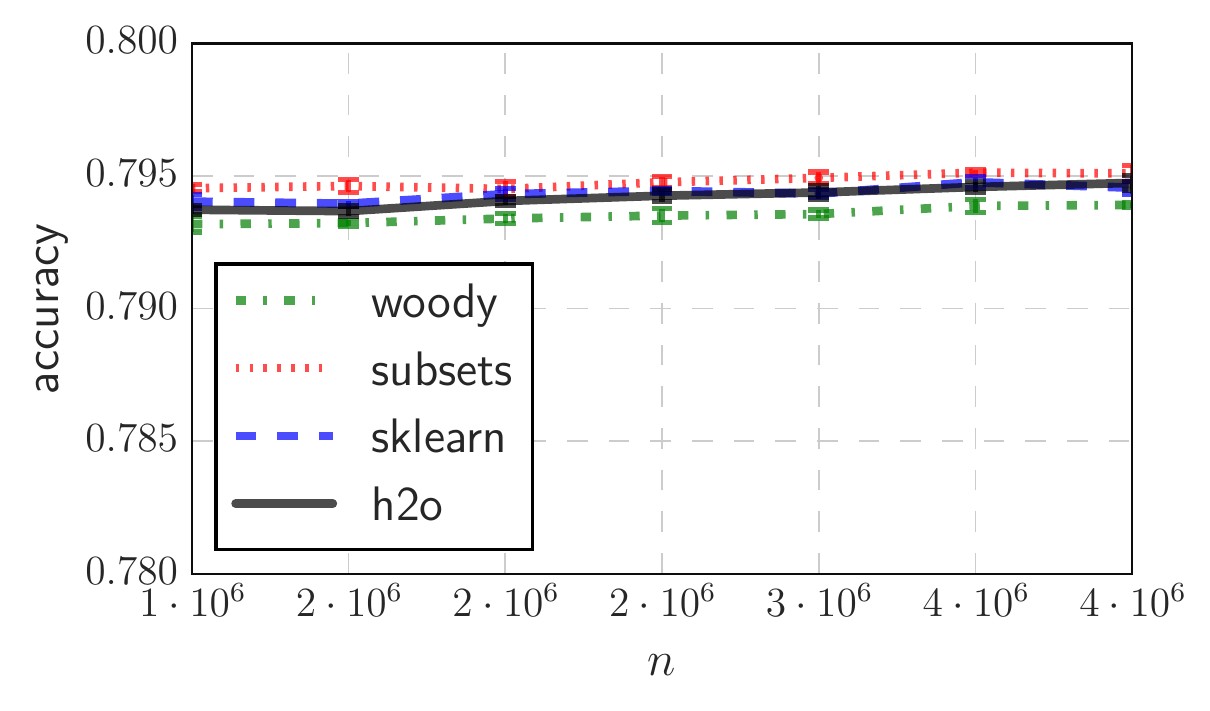}}
}
\subfigure[\higgs]{
\resizebox{0.31\textwidth}{!}{\includegraphics{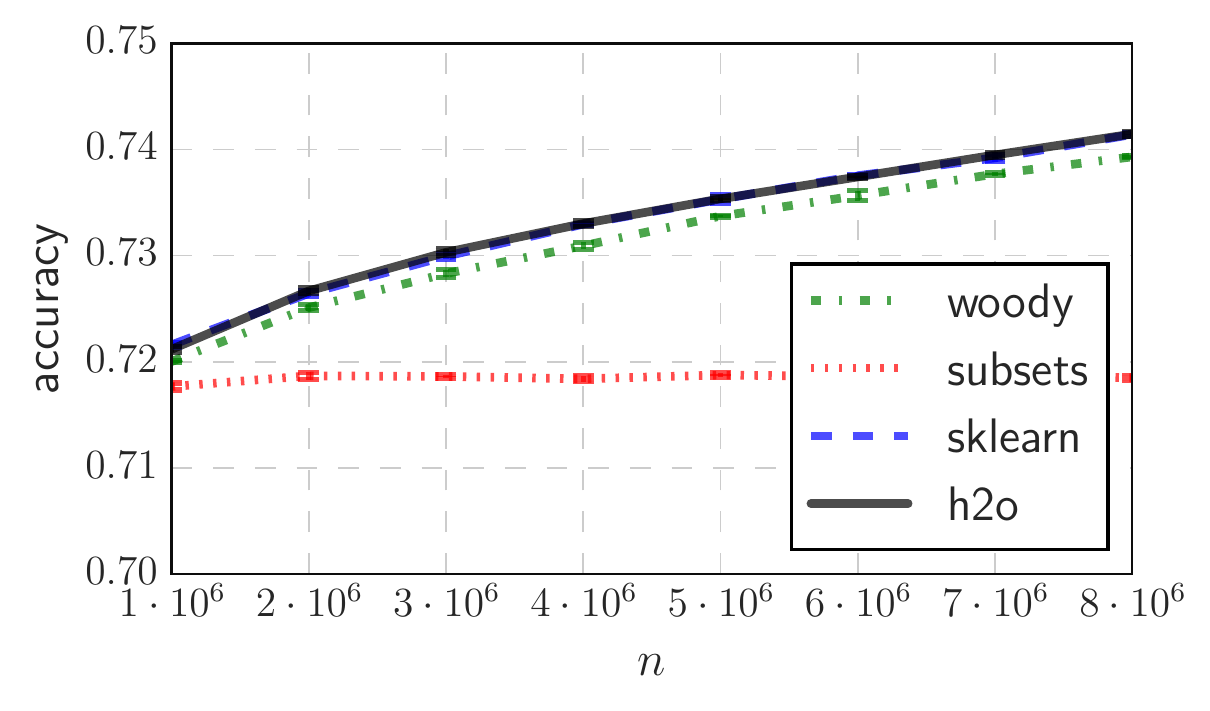}}
}
\vspace{-2ex}\caption{Training runtimes and test accuracies for the three medium-sized datasets (mean runtimes/accuracies as well as one standard deviations based on four runs with different seeds). The runtimes and accuracies are very similar to each other, indicating that the changes made for \woody do not significantly affect the outcome.
}
\label{fig:runtime_accuracies}
\end{figure*}
We consider two models to investigate the influence of $\lambda$: A standard random forest implementation that resorts to the adapted information gain criterion as well as~\woody. In both cases, we consider the \covtype dataset and different assignments for $\lambda$ ($\lambda=0,0.2,\ldots,1.0$). Both ensembles consider of 24 trees, where $\ntoptrees=6$ and $\nbottomtrees=4$ are used for \woody.

The outcome is shown in Figure~\ref{fig:influence_lambda}: As expected, smaller values for $\lambda$ generally yield slightly better accuracies. This is especially the case for the standard random forest implementation, which builds the full trees using the adapted criterion. However, the results also indicate that the adapted information gain does not severely affect the accuracy. Also, the accuracies still improve in general the more data are taken into account.\footnote{These results are in line with the ones reported for Mondarian Forests~\cite{LakshminarayananRT2014}, which resort to a label-independent information gain.} Finally, the influence of $\lambda$ is even less for \woody, which is due to the fact that the wrapper-based approach only resorts to $\bar{G}$ for the top trees. To conclude, we observe that the accuracy does not seem to be severely affected (see also below), especially in case $\bar{G}$ is used for the construction of the top trees only. However, balanced splits are important to quickly reduce the nodes' sizes for \woody. This is especially the case for almost pure nodes, which still have to be split to achieve the desired leaf sizes~$M$.

\subsection{Small Data}
Next, we compare the training times and test accuracies induced by the \covtype, \susy, and \higgs datasets. Again, we consider ensembles consisting of 24 classification trees ($\ntoptrees=6$ and $\nbottomtrees=4$ for \woody); all other parameters are set to the values described in Appendix~\ref{appendix:parameters}. The results are shown in Figure~\ref{fig:runtime_accuracies}: Except for \subsetwood, all implementations yield similar results, with \htwo~exhibiting a slightly worse classification accuracy on the \covtype~dataset (most likely due to the maximum tree depth of 20 not being sufficient for this dataset). The training runtimes are also similar among all implementations, with \woody being slightly faster than \sklearn on both the \susy~and the \higgs~dataset and slightly slower on the \covtype dataset. The \subsetwood scheme performs worse on both the \covtype and \higgs dataset. In both cases, it can be seen that not taking all the available training instances into account leads to a ``stagnating'' and slightly worse accuracy.\footnote{While \subsetwood performs slightly better on \susy, we do not consider the differences to be relevant (less than $0.2$\% differences w.r.t. the classification accuracy).}

To conclude, all implementations yield very similar accuracies and the modifications incorporated for \woody do not seem to severely affect the performance. However, the wrapper-based construction scheme can successfully incorporate significantly larger datasets, which potentially yields much better ensembles (see below). The direct competitor, \subsetwood, which only builds trees for subsets of instances, does not seem to benefit from more data, as it is indicated by the worse performance on both \covtype and \higgs. Finally, we would like to stress that \emph{fully-grown} trees are obtained via \woody, which might be a crucial for dealing with rare instances.





\subsection{Big Data}
\begin{figure}[t!]
\centering
\subfigure[Runtime]{
\resizebox{0.4\textwidth}{!}{\includegraphics{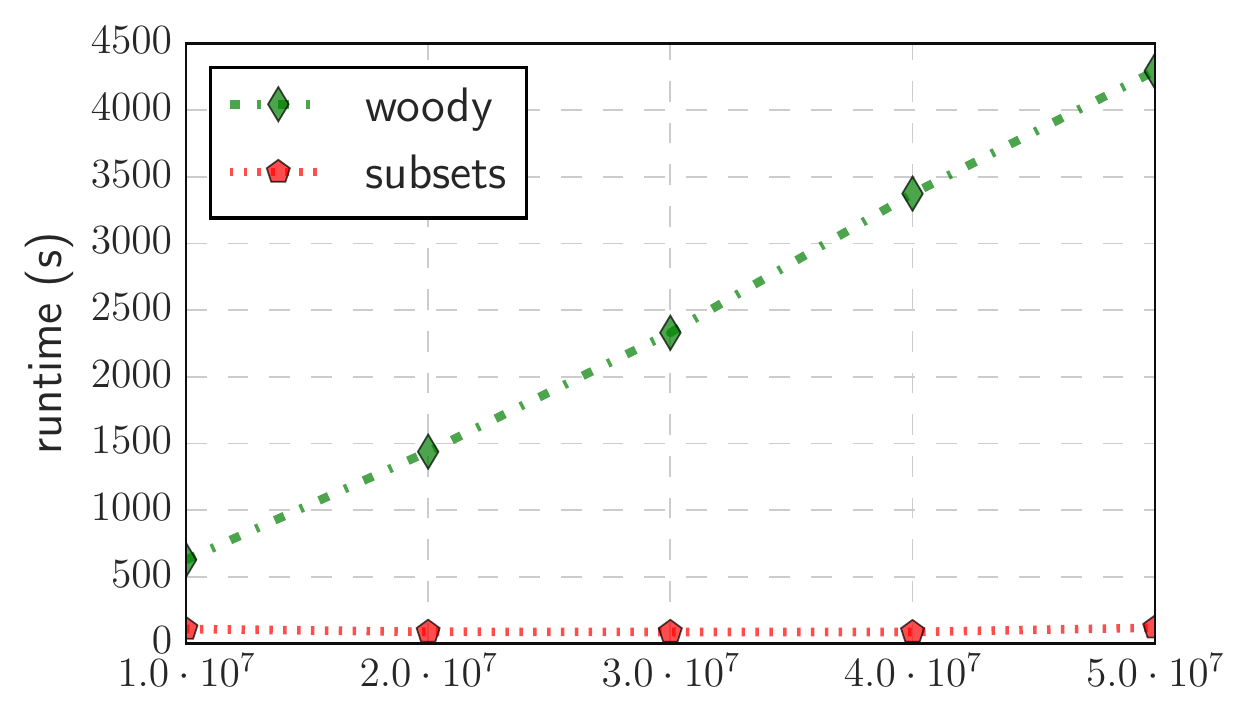}}
}
\subfigure[Accuracy]{
\resizebox{0.4\textwidth}{!}{\includegraphics{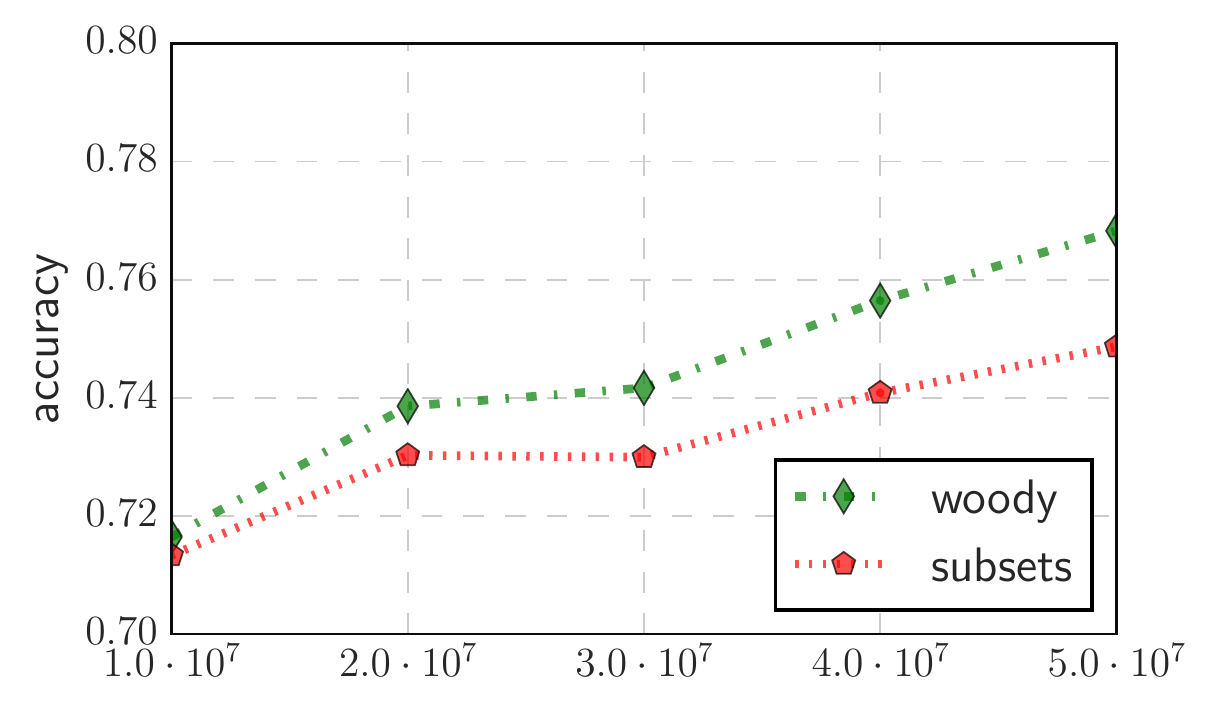}}
}
\vspace{-2ex}\caption{Training runtimes and test accuracies for the \landsat dataset with up to 50 million instances. The \sklearn implementation could not handle more than 10 million training instances due to memory errors. While the \htwo implementation could process the dataset instances, the accuracy on the test set was below 0.24 in all cases.}
\label{fig:runtime_accuracies_landsat}
\end{figure}
Next, we make use of the \landsat~dataset described in Appendix~\ref{appendix:landsat} and consider up to 50 million training instances; the
classification accuracies are evaluated on about three million test instances. We consider the same parameters for the different implementations as before and consider 12 estimators ($\ntoptrees=3$ and $\nbottomtrees=4$ as well as $\lambda=1.0$ for \woody). Intermediate results are now stored on disk instead in main memory for \woody and \subsetwood.

The results are shown in Figure~\ref{fig:runtime_accuracies_landsat}. It
can be seen that both \woody and \subsetwood are able to successfully process all training
instances. Further, taking more training instances into account leads to better
accuracies. The average accuracy of \subsetwood is also slightly worse than the one of \woody. The other two implementations could not handle these large-scale scenarios well: The \sklearn~implementation was only able to deal with ten million training instances without running into memory problems. For this single dataset instance, it yielded an average test accuracy of $0.72$, which was about the same as the one achieved by \woody in this case. While the \htwo implementation could process dataset instances with up to 30 million instances, the average test accuracy was below $0.24$ in all cases (not shown). To conclude, \woody is capable of taking all the available training instances into account. While the improvement over \subsetwood w.r.t. the accuracy might be moderate for the dataset at hand, we would like to stress \woody's capability to build full trees while taking all the training instances into account---which can be crucial for correctly classifying rare instances.

\begin{figure}[t]
\centering
\resizebox{0.4\textwidth}{!}{\includegraphics{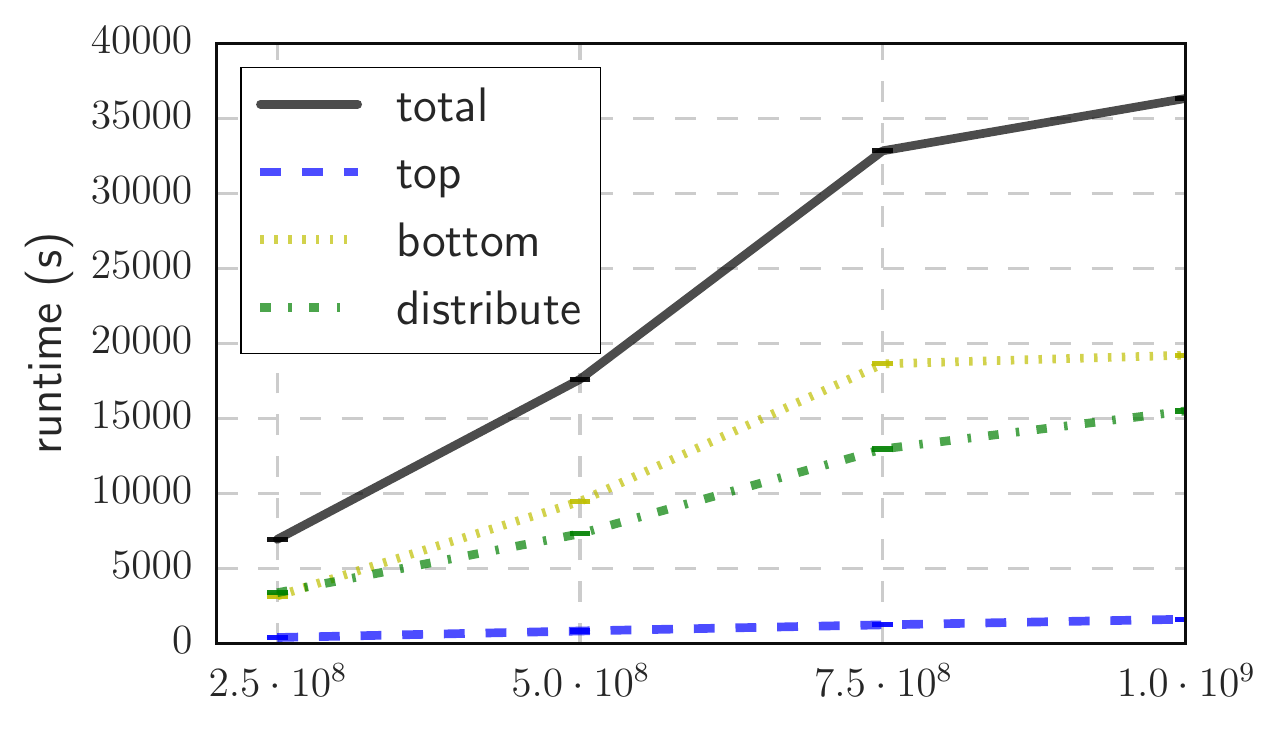}}
\vspace{-2ex}\caption{Runtimes for the different phases of \woody for the \landsat~dataset using up to one billion instances.}
\label{fig:runtime_large}
\end{figure}

Finally, we make use of all one billion training instances given in the \landsat~dataset and evaluate the runtime behavior of the \woody~implementation ($\ntoptrees=1$ and $\nbottomtrees=4$). This dataset contains very dominant classes (\eg, the 'water' class) and, thus, can lead to very unbalanced top trees in case the standard information gain criterion $G$ is used. However, using the modified information gain $\bar{G}$ with $\lambda=1$ yields balanced top trees with only few leaves (about 1,000). Hence, such top trees can be used to successfully partition huge datasets into smaller chunks. The runtimes of the different phases (single run) are shown in Figure~\ref{fig:runtime_large}. It can be seen that \woody can efficiently handle the one billion training instances with the distribution and bottom tree construction phases dominating the practical runtime.

\section{Conclusion}
\label{section:conclusion}
We propose \woody, a wrapper-based construction framework for building large
random forests for hundreds of millions of training instances. The key
idea is to use top trees to partition all the available training data
into smaller subsets associated with the top trees' leaves and to
build bottom trees for these subsets. While being conceptually simple,
the framework allows the construction of ensembles with fully-grown
trees for very large datasets. The practical benefits of our approach were empirically demonstrated on three medium-sized datasets and a large-scale application from the field of remote sensing with up to $10^9$ data points.

We expect these results to carry over to other learning tasks making \woody a powerful tool for mining big datasets---without requiring expensive compute resources. To the best of our knowledge, \woody is the first implementation that renders the construction of random forests possible for datasets containing hundreds of millions of instances using a standard desktop computer. We think that the \woody implementation---made publicly available on \url{https://github.com/gieseke/woody}---will be of significant practical importance for many real-world tasks in future.


\bibliographystyle{ACM-Reference-Format}
\bibliography{biblio}

\appendix
\begin{figure}[t]
\centering
\subfigure[\texttt{LC08\_L1TP\_196022\_20150415\_20170409\_01\_T1}]{
\mycolorbox{\resizebox{0.47\columnwidth}{!}{\includegraphics{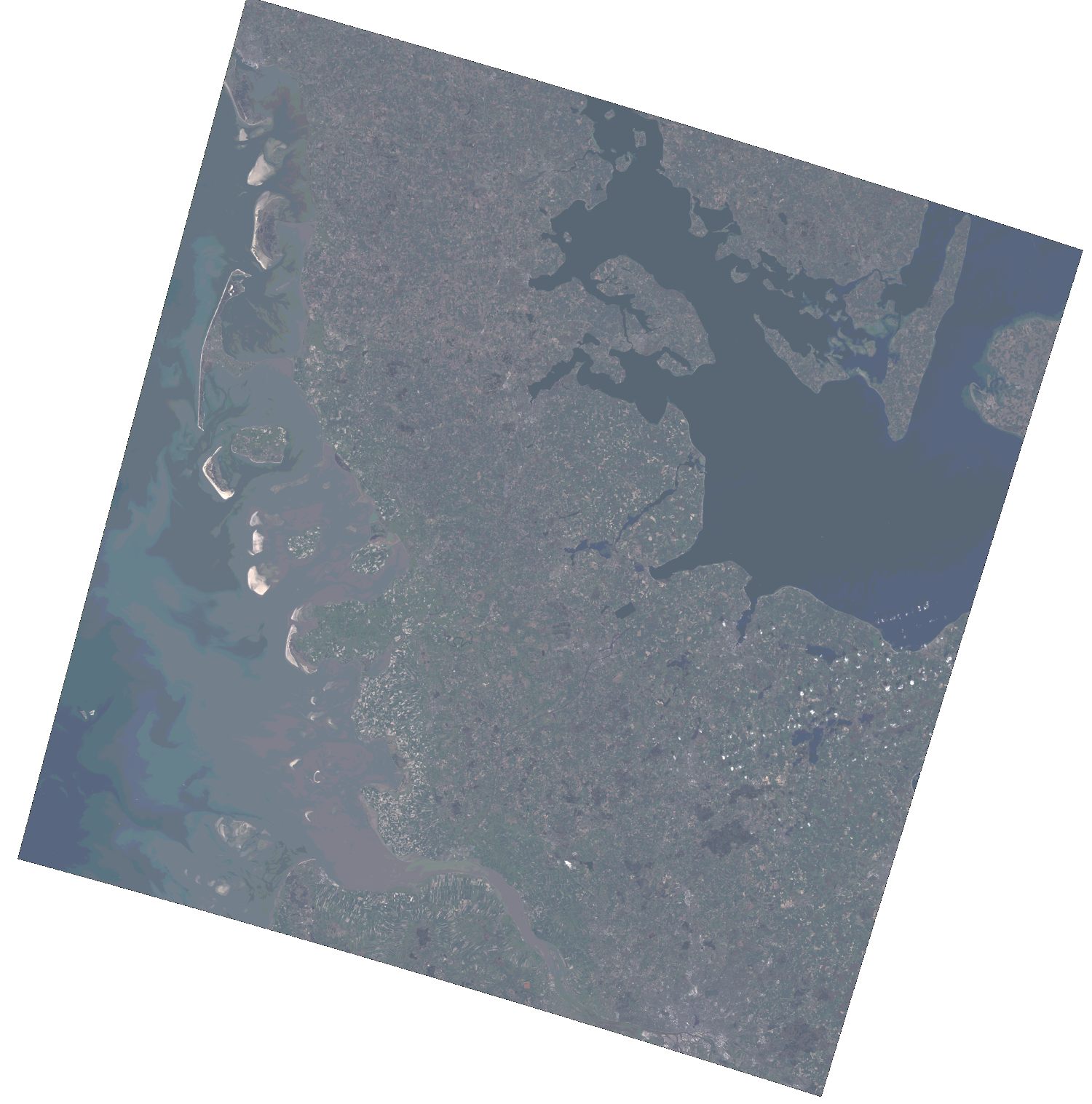}}}
\mycolorbox{\resizebox{0.47\columnwidth}{!}{\includegraphics{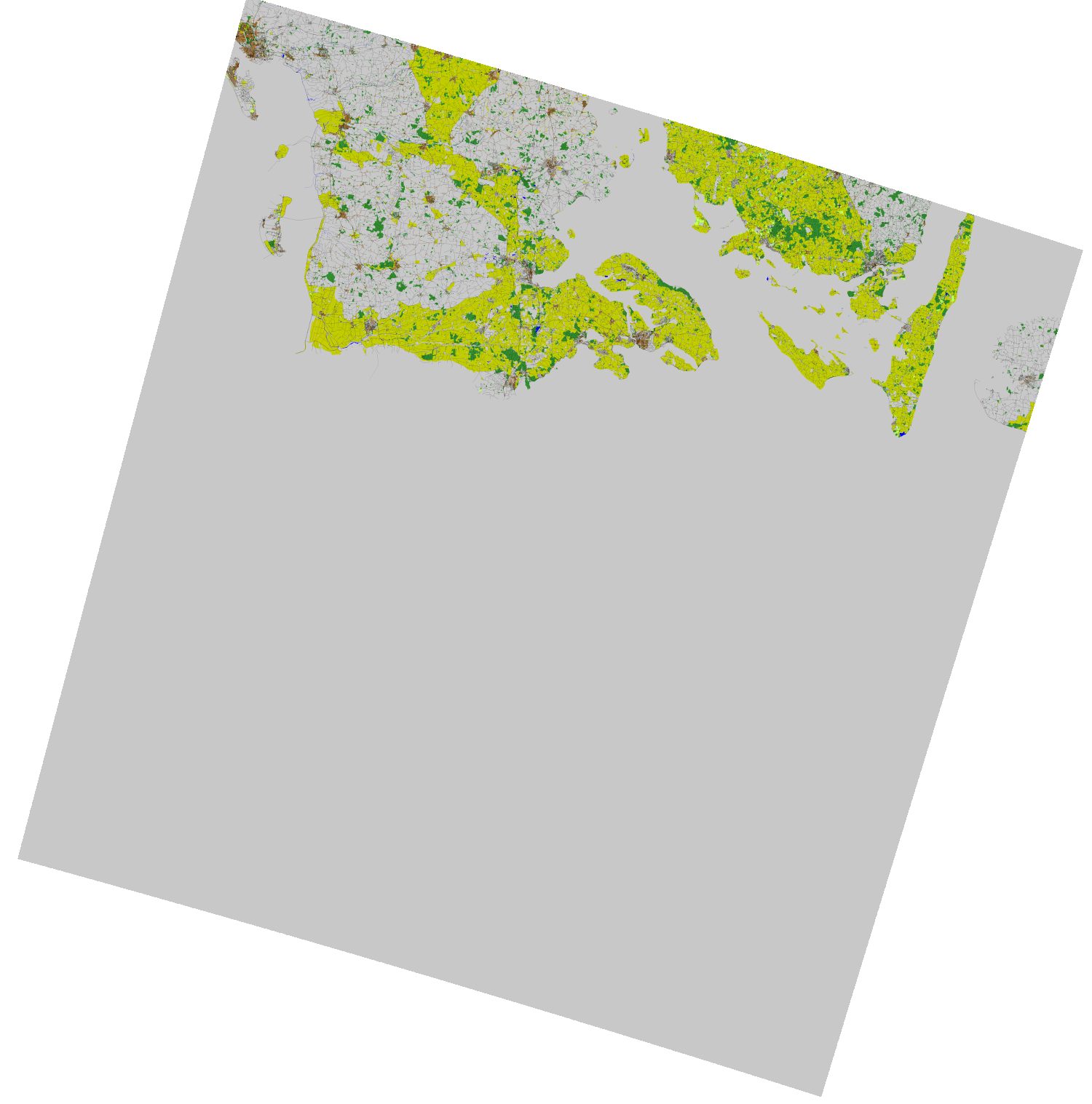}}}
}
\vspace{-2ex}\caption{Illustration of the first part of the \landsat dataset (used for the results shown in Figure~\ref{fig:runtime_accuracies_landsat}). The gray pixels are unlabeled data points.
}
\label{fig:landat_data}
\end{figure}

\section{Parameters}
\label{appendix:parameters}
If not stated otherwise, the parameters are fixed to the following values (using the same notation as for \sklearn): \texttt{max\_features="sqrt"}, \texttt{bootstrap=True}, \texttt{criterion="gini"}, \texttt{n\_jobs=4}, \texttt{max\_depth=None}, \texttt{min\_samples\_leaf=1}, and \texttt{min\_samples\_split=2}. Note that \woody uses \texttt{C} code for the construction of bottom trees, which follows the way random forests are built by \sklearn. The parameters for \htwo are adapted accordingly. In contrast to the other two implementations, the maximum tree depths is set to 20 for \htwo (default value); larger tree depths led to memory errors.

For \woody, the bottom trees are built in the same way as via the \sklearn implementation (using the same parameters), except for the \texttt{bootstrap} parameter (bootstrap samples are extracted during the distribution phase). Both, the number of samples for the top trees as well as the desired leaf sizes for the bottom trees are defined via $\min(500000, n, \max(100 \sqrt{\tsize}, 100000))$. For the experiments, either a \texttt{MemoryStore} or a \texttt{DiskStore} is considered. The former one is used for the runtime comparison shown in Figure~\ref{fig:runtime_accuracies}, where both the training data as well as the intermediate results are stored in memory (to obtain a fair comparison with \sklearn). For the other large-scale experiments, \texttt{DiskStore} is used that loads the data from disk (in chunks) and also stores the intermediate results on disk. For the \covtype dataset, the chunk size was fixed to $\chunksize=100,000$, whereas for all other datasets, a chunk size of $\chunksize=1,000,000$ was used. For \subsetwood, we consider subsets of size $50,000$ for \covtype and of size $500,000$ for all other datasets. We refer to \url{https://github.com/gieseke/woody} for the code and the experimental setup.

\section{Landsat-OSM}
\label{appendix:landsat}
The features for the \landsat~dataset are based on satellite data from the \emph{Landsat~8}~\cite{WULDER20122} project, see Figure~\ref{fig:landat_data}. The associated labels stem from the \emph{OpenStreetMap}~(OSM)~\cite{OpenStreetMap} project. More precisely, we consider $9$ bands (grayscale images) and $3 \times 3$ image patches, resulting in $81$ features. We extract such patches from the following Landsat scenes:\footnote{Downloadable via \url{https://earthexplorer.usgs.gov/}.} 
\begin{itemize}
 \item \texttt{LC08\_L1TP\_193022\_20170501\_20170515\_01\_T1}
 \item \texttt{LC08\_L1TP\_194022\_20160606\_20170324\_01\_T1}
 \item \texttt{LC08\_L1TP\_195021\_20160512\_20170324\_01\_T1}
 \item \texttt{LC08\_L1TP\_195022\_20160512\_20170324\_01\_T1}
 \item \texttt{LC08\_L1TP\_196021\_20150821\_20170406\_01\_T1}
 \item \texttt{LC08\_L1TP\_196022\_20150415\_20170409\_01\_T1}
 \item \texttt{LC08\_L1TP\_197020\_20150422\_20170409\_01\_T1}
 \item \texttt{LC08\_L1TP\_197021\_20150422\_20170409\_01\_T1}
\end{itemize}
The \texttt{LC08\_L1TP\_196022\_20150415\_20170409\_01\_T1} scene is split into two parts, where 5\% are used as test set (random subset) and the remaining 95\% as training set. The other scenes are attached to the training set for the runtime evaluation provided in Figure~\ref{fig:runtime_large}, yielding one billion training instances. Prior to extracting the patches, we pansparpened all images~\cite{WULDER20122}.

Finally, the label for each such image patch is based on an OSM label extracted for the pixel at the center of the patch. The following OSM labels are extracted for all patches~\cite{OpenStreetMap}:
 1: \texttt{landuse:forest},
 2: \texttt{landuse:meadow},
 3: \texttt{waterway:riverbank},
 4: \texttt{highway:all},
 5: \texttt{building:all},
 6: \texttt{landuse:reservoir},
 7: \texttt{natural:grassland},
 8: \texttt{railway:light\_rail}, and
 9: \texttt{landuse:farmland}.
The overall dataset consists of more than one billion labeled patches and provides a realistic classification benchmark scenario.

\end{document}